\documentclass[11pt,letterpaper]{article}
\usepackage[english]{babel}

\usepackage[margin=1in]{geometry}

\usepackage{natbib}

\usepackage[utf8]{inputenc} % allow utf-8 input
\usepackage[T1]{fontenc}    % use 8-bit T1 fonts
\usepackage{url}            % simple URL typesetting
\usepackage{booktabs}       % professional-quality tables
\usepackage{amsfonts}       % blackboard math symbols
\usepackage{nicefrac}       % compact symbols for 1/2, etc.
\usepackage{microtype}      % microtypography
\usepackage{xcolor}         % colors

\usepackage{amsmath}
\usepackage{amssymb}
\usepackage{mathtools}
\usepackage{amsthm}
\usepackage{thmtools, thm-restate}

\theoremstyle{plain}
\newtheorem{theorem}{Theorem}[section]

\newtheorem{lemma}[theorem]{Lemma}

\theoremstyle{definition}

\theoremstyle{remark}

\numberwithin{equation}{section}

\usepackage[unicode,psdextra,hypertexnames=false]{hyperref}
\hypersetup{
  colorlinks   = true, %Colours links instead of ugly boxes
  urlcolor     = blue, %Colour for external hyperlinks
  linkcolor    = blue, %Colour of internal links
  citecolor   = blue, %Colour of citations
  pdfauthor  = {},
  pdfsubject = {},
  pdfkeywords = {}
}
\usepackage{cleveref}
\usepackage{xspace}
\usepackage[vlined,linesnumbered,ruled]{algorithm2e}
\usepackage{float}
\usepackage{wrapfig}
\usepackage{multirow}
\usepackage{colortbl}
\usepackage{hhline}
\usepackage{physics2}
\usephysicsmodule{ab}      % 
\usephysicsmodule{diagmat} % 
\usephysicsmodule{braket}  % 
\usepackage{graphicx}
\usepackage{authblk}

\usepackage{fixdif,derivative}
\usepackage{bm}
\usepackage{bbm}
\usepackage{enumitem}

\renewcommand{\ge}{\geqslant}
\renewcommand{\le}{\leqslant}

\DeclareMathOperator*{\argmax}{arg\,max}

\newcommand{\procname}[1]{\ifmmode\mathop{\text{\normalfont\scshape #1}}\else{\normalfont\scshape #1}\xspace\fi}
\newcommand{\QMC}{\mathrm{QMC}}            % QMC algorithm
\newcommand{\QMCDesignEstimate}{\procname{QMCDesignEstimate}}       % QMC design estimate
\usepackage{array}
\newcommand{\SoftQME}{\procname{SoftQME}}
\newcommand{\SoftDesignEstimate}{\procname{SoftDesignEstimate}}
\newcommand{\SoftGElim}{\procname{Soft-G-Elim}}
\newcommand{\QGElim}{\procname{QG-Elim}}

\begin{document}

\title{Regret Bounds for Quantum Linear Bandits}
\author[1]{Maoli Liu}
\author[2]{Zhuohua Li}
\author[1]{John C.S. Lui}
\affil[1]{The Chinese University of Hong Kong}
\affil[2]{Xidian University}
\affil[ ]{ mlliu@cse.cuhk.edu.hk, lizhuohua@xidian.edu.cn, cslui@cse.cuhk.edu.hk}
\date{}

\maketitle

\begin{abstract}
We study \(d\)-dimensional quantum linear bandits with bounded rewards, where the learner selects an action and queries its quantum reward oracle or its inverse in each round.
For general action sets, existing algorithms achieve regret \(O(d^{3/2}\operatorname{polylog}(T))\) over \(T\) rounds.
Prior work established a lower bound of \(\Omega(d\log(1+T/d))\), leaving open whether the \(d^{3/2}\) dependence in the upper bound is necessary.
We resolve this question by proving a minimax lower bound of \(\Omega(d^{3/2}\log(1+T/d^{3/2}))\).
To prove this result, we construct a family of hard instances in which the reward gap of each action is proportional to its squared distance from the optimal action.
We establish a posterior Fisher information bound for adaptive quantum queries with quantum memory, relating the information acquired about the optimal direction to the regret incurred.
Viewing each selected action as an estimate of the optimal direction, we use a matrix Cram\'er--Rao argument to derive a lower bound on expected regret.
When the action set is fixed and has size \(K\), we propose two phased elimination algorithms, \SoftGElim and \QGElim.
We adapt quantum phase estimation to construct \SoftQME, a mean estimator with small bias and controlled tails.
\SoftGElim estimates the mean reward of every surviving action from independent \SoftQME outputs on the support of an experimental design.
Controlling the bias and fluctuations of these estimates yields regret \(O(d\log T\log(dT)\sqrt{\log(KT)})\), reducing the \(\log(KT)\log\log(dT)\) factor in the previous regret bound to \(\sqrt{\log(KT)}\). 
\QGElim further reduces this factor to \(\min\{\sqrt{\log(KT)},\sqrt d\}\).
\end{abstract}

\section{Introduction}
Stochastic linear bandits are a fundamental model for sequential decision making~\citep{lattimore2020bandit,bubeck2012regret,dani2008stochastic,abbasi2011improved}, with applications in personalized recommendation~\citep{li2010contextual}, dynamic pricing~\citep{cheung2019learning}, and network routing~\citep{awerbuch2004adaptive}.
Over \(T\) rounds, a learner repeatedly selects an action \(x\in\mathcal A\subseteq\mathbb R^d\) and observes a random reward with mean \(x^\top\theta\), where \(\theta\in\mathbb R^d\) is unknown.
The objective is to minimize cumulative regret, defined as the difference between the cumulative mean reward of always playing an optimal action and that of the learner~\citep{lattimore2020bandit}.
Regret bounds depend on the structure of the action set.
For general action sets, algorithms achieve regret of order \(d\sqrt T\) up to logarithmic factors~\citep{dani2008stochastic,abbasi2011improved}.
For finite action sets of size \(K\), regret bounds of order \(\sqrt{dT}\) are available up to logarithmic factors in \(K\) and \(T\)~\citep{chu2011contextual,li2019nearly,lattimore2020learning}.

Quantum linear bandits extend this model by giving the learner access to a quantum reward oracle and its inverse for each action~\citep{wan2023quantum}.
Quantum mean estimation provides a quadratic speedup over classical sampling for estimating bounded reward means~\citep{brassard2002quantum,montanaro2015quantum}.
This speedup enables quantum linear bandit algorithms to achieve regret bounds with polylogarithmic dependence on \(T\).
For general action sets, \citet{wan2023quantum} obtain a regret bound of \(O(d^2\log^{5/2}T)\).
Specializing the kernelized bandit analysis of~\citet{hikima2024quantum} to linear rewards yields a regret bound of \(O(d^{3/2}\log^{3/2}T\log(dT))\)~\citep{liu2026quantum}.
For finite action sets, \citet{liu2026quantum} combine quantum mean estimation with experimental design to achieve regret \(O(d\log T\log(KT)\log(dT)\log\log(dT))\), which has nearly linear dependence on \(d\) when \(K\) is polynomial in \(d\).

Despite these advances, the lower bound of \(\Omega(d\log(1+T/d))\) established by~\citet{liu2026quantum} leaves open whether the \(d^{3/2}\) dependence in the upper bound for general action sets is necessary.
For finite action sets, the optimal joint dependence of regret on \(d\), \(K\), and \(T\) also remains unresolved.
In this work, we resolve the question for general action sets and improve the regret upper bound for finite action sets.

\subsection{Results and Technical Overview}
\textbf{Lower bound for general action sets.}
For general action sets, Theorem~\ref{thm:general-lower} establishes the minimax regret lower bound
\[
    \Omega\left(d^{3/2}\log\left(1+\frac{T}{d^{3/2}}\right)\right)
\]
for all \(d\ge2\) and \(T\ge1\).
This lower bound matches the best known upper bound in its polynomial dependence on \(d\).

\citet{liu2026quantum} embed a \(d\)-armed bandit problem into the linear bandit setting, yielding the leading factor \(d\) in their \(\Omega(d\log(1+T/d))\) regret lower bound.
In their construction, each alternative changes only one arm, so the proof sums the costs of single-arm tests.
To establish the \(d^{3/2}\) dependence, we instead construct a family of instances with a spherical cap as the action set, where a single query can provide information about several parameter directions.
On these instances, the optimal action is \(\theta/\|\theta\|_2\), and the regret incurred by an action is proportional to its squared Euclidean distance from the optimal action.
Perturbations orthogonal to \(\theta\) change the optimal direction but have little effect on the mean rewards of actions near the optimum.
To turn this observation into a regret lower bound, we need to quantify how much information the learner can acquire about these perturbations relative to the regret incurred.

Standard change-of-measure arguments for classical linear bandit lower bounds analyze information gained from sequential reward observations~\citep[Chapter~24]{lattimore2020bandit}.
However, in our quantum setting, a learner can retain a quantum state across oracle queries, so the information calculations based on individual reward observations do not directly apply.
A key technical contribution is a posterior Fisher information bound for adaptive quantum queries with quantum memory (Lemma~\ref{lem:posterior-fisher}).
For affine reward means, it bounds the expected value of a measure of posterior Fisher information in fixed parameter directions by its prior value plus the contributions of the oracle queries.
Combining this bound with the geometry of our hard instances, we relate information about the optimal direction to the regret incurred in previous rounds.
A matrix Cram\'er--Rao argument~\citep{dembo1991information} then yields a lower bound on the mean squared error in estimating the optimal direction and hence on regret.

\textbf{Algorithms for fixed finite action sets.}
For fixed finite action sets of size \(K\), we propose two algorithms, \SoftGElim and \QGElim.
Theorem~\ref{thm:soft-elimination} shows that \SoftGElim achieves
regret \(O(d\log T\log(dT)\sqrt{\log(KT)})\).
This improves the \(K\)-dependent bound of~\citet{liu2026quantum}
by reducing the factor \(\log(KT)\log\log(dT)\) to
\(\sqrt{\log(KT)}\).
Theorem~\ref{thm:combined-elimination} further shows that
\QGElim achieves regret
\[
    \mathbb E[R(T)]
    =O\left(
        d\log T\log(dT)
        \min\left\{\sqrt{\log(KT)},\sqrt d\right\}
    \right).
\]
When \(K\) is polynomial in \(d\), the regret bound has nearly linear dependence on \(d\), whereas for \(K\) exponential in \(d\), its \(d^{3/2}\) dependence matches that of existing upper bounds for general action sets.
A similar distinction appears in classical linear bandits, with regret upper bounds of \(O(\sqrt{dT\log(KT)})\) for fixed finite action sets~\citep{lattimore2020learning} and \(\widetilde O(d\sqrt T)\) for general action sets~\citep{abbasi2011improved}.

To obtain these bounds, we reduce the query cost of estimating reward means within the experimental design and phased elimination framework of~\citet{liu2026quantum}.
In each phase, both algorithms estimate the reward means of \(O(d)\) actions in the support of a design and use weighted sums of these estimates to estimate the reward means of all surviving actions.
For \SoftGElim, we adapt the phase and probability estimation methods of~\citet{vanapeldoorn2023tomography} to obtain a reward mean estimator with small bias and controlled tails within a deterministic query budget, and apply it independently to each support action.
The bias guarantee keeps each weighted sum close to the true reward mean in expectation, while the tail guarantees and independence allow us to control its fluctuations through concentration.
We choose the estimation accuracy for each support action according to the design weights so that the reward means of all surviving actions are estimated accurately with high probability at a lower query cost.
Combining this query bound with phased elimination yields the \(\sqrt{\log(KT)}\) factor in the regret bound for \SoftGElim.
The algorithm \QGElim combines this estimation procedure with the Quantum Monte Carlo (QMC) procedure of~\citet[Lemma~5.6]{liu2026quantum}, whose query bound is independent of \(K\).
In each phase, \QGElim selects the procedure with the smaller query budget, yielding the minimum of the two terms in the regret bound.
Table~\ref{tab:regret-bounds} compares the regret bounds.

\begin{table}[t]
\centering
\caption{Regret bounds for quantum linear bandits with bounded rewards.}
\label{tab:regret-bounds}
\renewcommand{\arraystretch}{1.3}
\setlength{\tabcolsep}{4pt}
\begin{tabular}{
    >{\raggedright\arraybackslash}m{\dimexpr0.23\linewidth-2\tabcolsep\relax}
    >{\raggedright\arraybackslash}m{\dimexpr0.32\linewidth-2\tabcolsep\relax}
    >{\raggedright\arraybackslash}m{\dimexpr0.45\linewidth-2\tabcolsep\relax}}
\toprule
Reference & General action sets & Fixed finite action sets\\
\midrule
\citet{wan2023quantum}
   & \(O(d^2\log^{5/2}T)\) & \(O(d^2\log^{5/2}T)\) \\
\citet{hikima2024quantum}\(^{\dagger}\)
   & \(O(d^{3/2}\log^{3/2}T\log(dT))\)
   & \(O(d^{3/2}\log^{3/2}T\log(dT))\) \\
\multirow{2}{*}{\citet{liu2026quantum}} & ---
   & \(O(d\log T\log(KT)\log(dT)\log\log(dT))\) \\
 & --- & \(O(d^{3/2}\log T\log(dT))\) \\
\citet{liu2026quantum}
   & \(\Omega(d\log(1+T/d))\) & \(\Omega(d\log(1+T/d))\) \\
\rowcolor{gray!15}
\textbf{This work}
   & \(\Omega(d^{3/2}\log(1+T/d^{3/2}))\)\newline
     Theorem~\ref{thm:general-lower}
   & \(O(d\log T\log(dT)\min\{\sqrt{\log(KT)},\sqrt d\})\)\newline
     Theorem~\ref{thm:combined-elimination} \\
\bottomrule
\end{tabular}
\smallskip
\begin{minipage}{\linewidth}
\footnotesize
\raggedright
\(^{\dagger}\)\,The linear specialization is derived in~\citet[Appendix~C]{liu2026quantum}.
\end{minipage}
\end{table}

\subsection{Related Work}
\citet{wan2023quantum} study regret minimization in quantum multi-armed and linear bandits and propose algorithms with regret bounds of \(O(K\log T)\) for \(K\)-armed bandits and \(O(d^2\operatorname{polylog}(T))\) for \(d\)-dimensional linear bandits.
\citet{dai2023quantum} extend the quantum bandit framework to Bayesian optimization with kernelized rewards.
\citet{hikima2024quantum} further analyze the regret of quantum kernelized bandit algorithms using biased QMC estimates.
Other extensions address heavy-tailed rewards~\citep{wu2023quantum}, Lipschitz reward functions~\citep{yi2026quantum}, and bandits with knapsacks~\citep{su2025quantum}.
\citet{liu2026quantum} prove minimax regret lower bounds of \(\Omega(K\log(1+T/K))\) for quantum multi-armed bandits and \(\Omega(d\log(1+T/d))\) for quantum linear bandits with general action sets.
To prove these bounds, they use the polynomial method to establish a lower bound on the query complexity of testing a fixed reward mean against an interval of alternatives.
A reduction from bandit regret to this testing problem then yields the regret bounds.
We strengthen their linear bandit lower bound by combining a posterior Fisher information bound with the geometry of a spherical cap to relate information acquisition to regret.

Classical linear bandits have been extensively studied in the literature~\citep{bubeck2012regret,lattimore2020bandit}.
For general, possibly infinite, action sets, \citet{dani2008stochastic,abbasi2011improved} develop algorithms based on the principle of optimism in the face of uncertainty, with regret of order \(d\sqrt T\) up to logarithmic factors.
\citet{chu2011contextual} propose SupLinUCB for linear contextual bandits with oblivious action sets of size \(K\), achieving regret of order \(\sqrt{dT}\) up to logarithmic factors in \(K\) and \(T\).
\citet{li2019nearly} refine this approach using variable confidence levels to reduce the logarithmic factors.
\citet{jin2026nearly} further studies arbitrary adaptive action sets and establishes nearly optimal regret bounds.
For fixed finite action sets, phased elimination algorithms~\citep{lattimore2020bandit,lattimore2020learning} use \(G\)-optimal designs~\citep{kiefer1960equivalence} to allocate samples across the surviving actions.

Quantum mean estimation builds on quantum amplitude estimation~\citep{brassard2002quantum}, which provides a quadratic speedup over classical sampling for estimating the means of bounded random variables.
\citet{montanaro2015quantum} develops quantum Monte Carlo algorithms for estimating the means of random variables with a known variance bound.
\citet{hamoudi2021quantum} develops a quantum mean estimator with sub-Gaussian error guarantees, while \citet{kothari2023mean} achieve optimal variance-dependent query complexity without requiring prior knowledge of the variance.
For vector-valued random variables, \citet{cornelissen2022near} develop nearly optimal quantum mean estimation algorithms with covariance-dependent error bounds.
\citet{cornelissen2023sublinear} develop nearly unbiased quantum amplitude and mean estimators with low variance.
\citet{liu2026quantum} adapt this amplitude estimator and combine it with experimental design to estimate reward means in quantum linear bandits with finite action sets.
\citet{vanapeldoorn2023tomography} develop phase estimation methods with small bias and high-probability accuracy guarantees and apply them to probability estimation.
We adapt these phase and probability estimation methods to obtain reward mean estimates with small bias and controlled tails within a deterministic query budget.

Quantum pure exploration focuses on identifying an optimal arm with as few oracle queries as possible.
\citet{wang2021quantum} obtain a quadratic speedup in query complexity up to logarithmic factors using a quantum oracle that allows arms to be queried in superposition.
\citet{wang2025best} further study how the query complexity depends on the number of arms that can be queried in superposition.
\citet{buchholz2025multi} study quantum channel oracles that restrict access to reward randomness and show that best arm identification can have the same query complexity as in the classical setting, even when arms can be queried in superposition.
In quantum optimization, coherent function evaluations enable quantum gradient estimation~\citep{jordan2005fast,gilyen2019optimizing}.
\citet{li2022quantum} develop quantum algorithms for approximately convex optimization and use them to obtain regret bounds with polylogarithmic dependence on \(T\) for stochastic convex bandits.
Their model permits queries at points in superposition and evaluates regret at a point reported after each query, whereas our model charges each query the gap of the classically selected action.
Another line of work studies bandits in which actions correspond to measurements of an unknown quantum state~\citep{lumbreras2022multi}.
For pure states, measurement noise can decrease as the chosen measurement approaches the optimum.
Motivated by this behavior, \citet{lumbreras2024linear} study linear bandits with vanishing noise and establish polylogarithmic minimax regret bounds.
\citet{lumbreras2026learning} develop protocols for pure state tomography with polylogarithmic regret and relate this regret to the cumulative disturbance caused by measurements.

\section{Preliminaries}
\label{sec:preliminaries}
\paragraph{Quantum linear bandits.}
We consider a quantum linear bandit problem with a known compact action set \(\mathcal A\subseteq\mathbb R^d\) and an unknown parameter \(\theta\in\mathbb R^d\).
For each action \(x\in\mathcal A\), the mean reward is \(\mu(x):=x^\top\theta\in[0,1]\).
When \(\mathcal A\) is finite, we write \(K:=|\mathcal A|\).

For each action \(x\), the learner has access to a quantum reward oracle \(\mathcal O_x\), a unitary satisfying
\begin{equation}
\label{eq:reward-oracle}
\mathcal O_x|0\rangle = \sum_{\omega\in\Omega_x}\sqrt{P_x(\omega)}\,|\omega\rangle|y_x(\omega)\rangle,
\end{equation}
where \(P_x\) is a probability distribution on \(\Omega_x\subseteq\Omega\) for a common finite set \(\Omega\).
The reward function \(y_x:\Omega_x\to[0,1]\) satisfies \(\mathbb E_{\omega\sim P_x}[y_x(\omega)]=\mu(x)\).
Here \(|0\rangle\) denotes the all-zero state of both registers.
Measuring the reward register of \(\mathcal O_x|0\rangle\) gives a classical reward sample for action \(x\).

A policy interacts with these oracles over a horizon of \(T\) rounds.
In round \(t\in\{1,2,\dots,T\}\), it selects an action \(x_t\in\mathcal A\) and makes at most one query to \(\mathcal O_{x_t}\) or \(\mathcal O_{x_t}^{\dagger}\).
The policy may retain finite-dimensional quantum memory across rounds and perform quantum operations and measurements between oracle queries.
These operations and action choices may depend on previous measurement outcomes and internal randomness.

The learner's objective is to maximize its expected cumulative reward, or equivalently, to minimize its expected cumulative regret \(\mathbb E[R(T)]\).
Let \(x^\star\in\argmax_{x\in\mathcal A}\mu(x)\) be an optimal action and define the reward gap \(\Delta(x):=\mu(x^\star)-\mu(x)\).
The policy incurs regret \(\Delta(x_t)\) in round \(t\), so the cumulative regret is
\[
    R(T):=\sum_{t=1}^T\Delta(x_t)
    =\sum_{t=1}^T(x^\star-x_t)^\top\theta.
\]

\paragraph{Quantum mean estimation.}
The following quantum Monte Carlo (QMC) guarantee bounds the number of oracle queries needed to estimate an action's mean reward to a prescribed accuracy.

\begin{lemma}[Quantum Monte Carlo {\citep[Theorem~2.3]{montanaro2015quantum}}]
\label{lem:qmc}
Let \(\mathcal O\) be a quantum reward oracle of the form~\eqref{eq:reward-oracle} with mean reward \(\mu\).
For any \(\epsilon\in(0,1]\) and \(\delta\in(0,1/2]\), the procedure \(\QMC(\mathcal O,\epsilon,\delta)\) outputs an estimate \(\widehat\mu\in[0,1]\) satisfying \(\Pr[|\widehat\mu-\mu|>\epsilon]\le\delta\).
It uses at most \(C_{\rm QMC}\epsilon^{-1}\log(1/\delta)\) queries to \(\mathcal O\) and \(\mathcal O^\dagger\), for a universal constant \(C_{\rm QMC}>0\).
\end{lemma}

\section{Lower Bound for General Action Sets}
\label{sec:general-lower}
We prove the following minimax regret lower bound for general action sets.

\begin{theorem}
\label{thm:general-lower}
There exists a universal constant \(c>0\) such that, for every \(d\ge2\), there is a compact subset \(\mathcal A\) of the unit sphere in \(\mathbb R^d\) with the following property.
For every horizon \(T\ge1\) and every policy on \(\mathcal A\) over \(T\) rounds, there exist a parameter \(\theta\in\mathbb R^d\) and corresponding quantum reward oracles for which
\[
    \mathbb E[R(T)]
    \ge c\,d^{3/2}\log\left(1+\frac{T}{d^{3/2}}\right).
\]
\end{theorem}

\paragraph{Proof intuition.}
The lower bound of~\citet{liu2026quantum} embeds a multi-armed bandit construction into a linear bandit with \(d\) actions.
In that construction, each alternative changes only one arm, and each query to that arm incurs constant regret on a common baseline instance.
The proof sums the costs of single-arm tests to obtain linear dependence on \(d\).

We use a spherical cap of actions and analyze regret through estimation of the optimal direction.
We choose a prior for which the optimal action is \(\theta/\|\theta\|_2\).
Since all actions have unit norm, each action's gap is proportional to its squared distance from this direction.
A single query on the cap can provide information about several parameter directions, so summing the costs of separate tests for these directions could count the same query more than once.
Since actions can have arbitrarily small gaps, we also need to bound their information contributions in terms of the regret incurred.
An action near the optimum has a small component orthogonal to \(\theta\), so its reward mean is weakly sensitive to parameter perturbations in those directions.

Classical lower bounds for the unit ball decompose the Kullback--Leibler (KL) divergence of the interaction history into contributions from individual reward observations~\citep[Section~24.2]{lattimore2020bandit}.
With quantum memory, a measurement can depend on several coherent oracle queries, so these contributions do not describe the information acquired by a quantum policy.
We establish a posterior Fisher information bound that retains the directions of the queried actions while allowing adaptive queries and quantum memory.

The bound applies to fixed parameter directions, whereas the directions orthogonal to the optimal action depend on the unknown parameter.
We therefore introduce an auxiliary noisy observation of \(\theta\) and use it to construct a reference direction close to the optimal action.
This observation is used only in the analysis and is not revealed to the policy.
Conditional on this observation, the reference direction is fixed, allowing us to bound posterior information in directions orthogonal to it.
Each query contributes according to the component of the selected action orthogonal to the reference direction.
The squared norm of this component is controlled by the action's gap and the reference error.
A more accurate auxiliary observation reduces the reference error but increases the Fisher information of the conditional prior.
Combining this information bound with a matrix Cram\'er--Rao argument gives a lower bound on the mean squared error of the selected action as an estimate of the optimal direction.
Choosing the accuracy of the auxiliary observation and using the relation between squared distance and regret yields an inequality linking current regret to past regret.
A weighted summation gives a lower bound on cumulative regret averaged over the prior, and hence the minimax lower bound.

We first state the posterior Fisher information bound for affine mean functions of an \(m\)-dimensional parameter.
For a compactly supported probability density \(p\) on \(\mathbb R^m\) whose square root has a square-integrable weak gradient, and a matrix \(B\in\mathbb R^{m\times k}\) with \(B^\top B=I_k\), define
\begin{equation}
\label{eq:posterior-fisher-definition}
    J(p):=4\int_{\mathbb R^m}
        \nabla\sqrt{p(z)}\,\nabla\sqrt{p(z)}^\top\,dz,
    \qquad
    S_B(p):=\operatorname{tr}\sqrt{B^\top J(p)B}.
\end{equation}
Here \(\nabla\sqrt p\) is the weak gradient, \(\operatorname{tr}\) denotes the trace, and the matrix square root is the positive semidefinite one.
Equivalently, \(J(p)=\int_{\{p>0\}}\nabla p(z)\,\nabla p(z)^\top/p(z)\,dz\), the usual Fisher information matrix.
The matrix \(B^\top J(p)B\) measures information in the directions spanned by the columns of \(B\).
The following lemma bounds the expected posterior value of \(S_B\) by its prior value plus the contributions of the oracle queries.

\begin{lemma}
\label{lem:posterior-fisher}
Let \(Z\in\mathbb R^m\) have a bounded, compactly supported probability density \(p\) whose square root has a square-integrable weak gradient.
Fix \(B\in\mathbb R^{m\times k}\) with \(B^\top B=I_k\).
For a realization \(Z=z\), each action \(x\) has mean reward \(b_x+a_x^\top z\) and reward oracle \(\mathcal O_x=U_{b_x+a_x^\top z}\).
The coefficients \(b_x\in\mathbb R\) and \(a_x\in\mathbb R^m\) depend measurably on \(x\) and satisfy \(\sup_x\|a_x\|_2<\infty\).
Assume that \(\mu\mapsto U_\mu\) is continuously differentiable on an open interval containing \(b_x+a_x^\top z\) for every action \(x\) and every \(z\) in some neighborhood of \(\operatorname{supp}(p)\).
Its derivative has operator norm \(\|\partial_\mu U_\mu\|_{\mathrm{op}}\le L\) throughout this interval.
For an integer \(q\ge0\), consider a protocol that selects an action \(x_s\) in each round \(s=1,\ldots,q\) and makes at most one query to \(\mathcal O_{x_s}\) or \(\mathcal O_{x_s}^{\dagger}\).
The protocol may retain finite-dimensional quantum memory across rounds and perform quantum operations and measurements between queries.
Its operations between queries and action choices may depend measurably on previous measurement outcomes and internal randomness, but not otherwise on \(Z\).
There exists a refinement \(Y\) of the protocol's classical record whose construction depends only on the protocol and not on the prior \(p\) or the matrix \(B\).
For every \(p\) and \(B\) satisfying the assumptions above, the posterior density \(p_Y\) of \(Z\) given \(Y\) has a square root with a square-integrable weak gradient almost surely, and
\begin{equation}
\label{eq:posterior-fisher-bound}
    \mathbb E_Y S_B(p_Y)
    \le S_B(p)+2L\sum_{s=1}^q
        \left(\mathbb E\|B^\top a_{x_s}\|_2^2\right)^{1/2}.
\end{equation}
\end{lemma}

The refined record \(Y\) augments the original record with additional information used only in the analysis and not revealed to the policy.
In~\eqref{eq:posterior-fisher-bound}, \(x_s\) is the action selected in round \(s\), and we set its coefficient to zero if that round contains no query.
All expectations use the given prior and protocol.
The proof exploits the fact that, for each refined record, each query contributes a term of rank at most one to the parameter derivative, avoiding an additional dimension factor in~\eqref{eq:posterior-fisher-bound}.
Appendix~\ref{app:posterior-fisher-proof} gives the proof, including continuous measurement outcomes.

\begin{proof}[Proof of Theorem~\ref{thm:general-lower}]
Let \(w=1/(8\sqrt d)\) and draw \(\theta=\tfrac12e_1+wZ\), where \(Z_1,\ldots,Z_d\) are independent with density \(\varphi(z)=\cos^2(\pi z/2)\mathbf1\{|z|\le1\}\).
Write \(p_0\) for the density of \(\theta\) and \(\mathcal Q=\frac12e_1+[-w,w]^d\) for its support.
Take the action set \(\mathcal A:=\{x\in\mathbb R^d:\|x\|_2=1,\ x_1\ge\sqrt3/2\}\), a spherical cap around \(e_1\).
Every parameter in \(\mathcal Q\) satisfies \(3/8\le\|\theta\|_2\le5/8\) and has optimal action \(g(\theta):=\theta/\|\theta\|_2\in\mathcal A\).
All action means lie in \([1/4,3/4]\).
We use Bernoulli reward oracles whose derivatives with respect to the mean have operator norm at most \(2/\sqrt3\).
Appendix~\ref{app:cap-instance} gives their construction and verifies these properties.
Since \(\|x\|_2=\|g(\theta)\|_2=1\), the gap of every \(x\in\mathcal A\) satisfies
\begin{equation}
\label{eq:cap-gap}
    \Delta_\theta(x)
    =\|\theta\|_2-x^\top\theta
    =\frac{\|\theta\|_2}{2}\|x-g(\theta)\|_2^2.
\end{equation}
Fix the policy and define \(r_t:=\mathbb E[\Delta_\theta(x_t)]\).
Write \(F_t:=\sum_{s=1}^t\sqrt{r_s}\), with \(F_0:=0\), to track the query contributions.
All expectations below include the prior and policy, as well as the auxiliary noise when present.
The constants \(c,C>0\) are universal and may change between inequalities.

We introduce an auxiliary observation to choose a reference direction close to \(g(\theta)\).
For an error scale \(\rho\in(0,1]\), let \(W=\theta+N\), where \(N\sim\mathcal N(0,\rho^2I_d/d)\) is independent of \(\theta\) and all randomness of the policy.
The policy does not observe \(W\).
Let \(v\) be the normalized Euclidean projection of \(W\) onto \(\mathcal Q\), and choose an orthonormal basis matrix \(B=B(W)\) for \(v^\perp\) measurably.
Conditional on \(W\), the matrix \(B\) is fixed and the density of \(\theta\) is
\[
    p_W(\theta)
    =\frac{p_0(\theta)\exp[-d\|W-\theta\|_2^2/(2\rho^2)]}
        {\int p_0(\vartheta)\exp[-d\|W-\vartheta\|_2^2/(2\rho^2)]\,d\vartheta}.
\]
The denominator is positive and finite for every \(W\).
The square root of \(p_0\) is Lipschitz and compactly supported, so multiplication by the smooth Gaussian factor shows that \(p_W\) is bounded and its square root has a square-integrable weak gradient.
Its score on \(\{p_0>0\}\) is \(\nabla\log p_W=\nabla\log p_0+d(W-\theta)/\rho^2\).
Since \(N\) is independent of \(\theta\) and has mean zero, the cross term vanishes after averaging, giving
\[
    \mathbb E_W\operatorname{tr}J(p_W)
    =\operatorname{tr}J(p_0)+\frac{d^2}{\rho^2}
    =64\pi^2d^2+\frac{d^2}{\rho^2}.
\]
Appendix~\ref{app:cap-instance} verifies the regularity and Fisher information of \(p_0\).
The inequality \(S_B(p_W)^2\le(d-1)\operatorname{tr}J(p_W)\) and Cauchy--Schwarz imply
\begin{equation}
\label{eq:localized-prior-fisher}
    \mathbb E_W S_{B(W)}(p_W)\le C\frac{d^{3/2}}\rho.
\end{equation}

Since the reward mean of action \(x\) is \(x^\top\theta\), the coefficients in Lemma~\ref{lem:posterior-fisher} are \(b_x=0\) and \(a_x=x\).
The projected coefficient is therefore \(B^\top x\).
Projection onto \(\mathcal Q\) does not increase the distance to \(\theta\), and the normalization bound in Appendix~\ref{app:cap-instance} gives \(\|v-g(\theta)\|_2\le(16/3)\|W-\theta\|_2\).
Since \(B^\top v=0\) and \(\|\theta\|_2\ge3/8\), equation~\eqref{eq:cap-gap} gives
\[
    \begin{aligned}
        \|B^\top x\|_2^2
        &\le 2\|x-g(\theta)\|_2^2+2\|g(\theta)-v\|_2^2
        \le \frac{32}{3}\Delta_\theta(x)+\frac{512}{9}\|W-\theta\|_2^2
        \le C\left(\Delta_\theta(x)+\|W-\theta\|_2^2\right).
    \end{aligned}
\]
Taking expectations yields
\begin{equation}
\label{eq:projected-action-regret}
    \mathbb E\|B^\top x_s\|_2^2\le C(r_s+\rho^2).
\end{equation}

We now apply Lemma~\ref{lem:posterior-fisher} conditional on \(W\), with the fixed matrix \(B(W)\), to the first \(t-1\) rounds and the subsequent choice of \(x_t\).
The lemma provides a refinement \(Y\) whose construction is independent of \(W\), with \(x_t\) included as the final classical output.
Since the policy does not observe \(W\), the likelihood of \(Y\) given \((\theta,W)\) equals its likelihood given \(\theta\).
Let \(p_{W,Y}\) be the posterior density of \(\theta\) given \(W,Y\).
For each query contribution, Jensen's inequality for the square-root function gives
\[
    \mathbb E_W\!\left[
        \sqrt{\mathbb E[\|B(W)^\top x_s\|_2^2\mid W]}
    \right]
    \le\sqrt{\mathbb E\|B(W)^\top x_s\|_2^2}.
\]
Averaging the lemma over \(W\) and using this inequality together with~\eqref{eq:localized-prior-fisher} and~\eqref{eq:projected-action-regret} gives
\begin{align}
    \mathbb E_{W,Y}S_{B(W)}(p_{W,Y})
    \le C\frac{d^{3/2}}\rho + C\sum_{s<t}\sqrt{r_s+\rho^2}
    \le C\left(\frac{d^{3/2}}\rho+F_{t-1}+t\rho\right).
    \label{eq:posterior-fisher-regret}
\end{align}

To convert this information bound into a regret bound, we regard \(x_t\) as an estimate of \(g(\theta)\).
Both \(v\) and \(g(\theta)\) lie in \(\mathcal A\), so their angle is at most \(\pi/3\) and \(v^\top g(\theta)\ge1/2\).
The derivative of \(g\) therefore satisfies
\[
    B^\top Dg(\theta)B
    =\frac{I_{d-1}-(B^\top g(\theta))(B^\top g(\theta))^\top}{\|\theta\|_2}
    \succeq\frac25 I_{d-1}.
\]
For any density \(p\) supported on \(\mathcal Q\) whose square root has a square-integrable weak gradient, this derivative bound and a matrix Cram\'er--Rao argument give, for each fixed \(x\in\mathcal A\),
\begin{equation}
\label{eq:posterior-direction-error}
    \mathbb E_p\|B^\top(x-g(\theta))\|_2^2
    \ge\frac4{25}\frac{(d-1)^3}{S_B(p)^2}.
\end{equation}
Appendix~\ref{app:direction-estimation} proves~\eqref{eq:posterior-direction-error} by weak integration by parts and matrix Cauchy--Schwarz.
The factor \((d-1)^3\) follows from H\"older's inequality, which bounds \(\operatorname{tr}[(B^\top J(p)B)^{-1}]\) below by \((d-1)^3/S_B(p)^2\).

Conditional on \(W,Y\), the action \(x_t\) is fixed and the posterior satisfies the required regularity.
Since \(\Delta_\theta(x_t)\ge(3/16)\|B^\top(x_t-g(\theta))\|_2^2\), we obtain
\[
    \mathbb E[\Delta_\theta(x_t)\mid W,Y]
    \ge\frac3{100}\frac{(d-1)^3}{S_{B(W)}(p_{W,Y})^2}.
\]
Averaging, applying Jensen's inequality to \(s\mapsto s^{-2}\), and using~\eqref{eq:posterior-fisher-regret} yields
\begin{equation}
\label{eq:localized-regret-recurrence}
    r_t
    \ge\frac{c(d-1)^3}{\bigl(\mathbb E_{W,Y}S_{B(W)}(p_{W,Y})\bigr)^2}
    \ge\frac{c\,d^3}
        {\left(d^{3/2}/\rho+F_{t-1}+t\rho\right)^2}.
\end{equation}
The last inequality uses \((d-1)^3\ge d^3/8\) for \(d\ge2\).

We finish by summing~\eqref{eq:localized-regret-recurrence} with suitable weights.
Set \(D:=d^{3/2}\) and \(u_t:=\sqrt{r_t}\).
Since the policy does not observe \(W\), varying \(\rho\) leaves \(r_t\) unchanged.
Choosing \(\rho=\sqrt{D/(t+D)}\) for each \(t\) gives
\begin{equation}
\label{eq:root-regret-recurrence}
    u_t\left(C\sqrt{D(t+D)}+F_{t-1}\right)\ge cD.
\end{equation}
To obtain the cumulative regret lower bound, we sum~\eqref{eq:root-regret-recurrence} with weights \(1/(t+D)\).
The main step is to bound the cross term \(\sum_{t=1}^T u_tF_{t-1}/(t+D)\) by a constant multiple of \(\mathbb E[R(T)]\).
For \(s<t\), we use
\[
    \frac{2u_su_t}{t}
    \le\frac{\sqrt{s}}{t^{3/2}}u_s^2
        +\frac{1}{\sqrt{st}}u_t^2.
\]
Summing this inequality and using \(\sum_{t>s}t^{-3/2}\le2/\sqrt{s}\) and \(\sum_{s<t}s^{-1/2}\le2\sqrt{t}\) yields
\begin{equation}
\label{eq:weighted-regret-sum}
    \sum_{t=1}^T\frac{u_tF_{t-1}}{t+D}
    \le\sum_{s<t}\frac{u_su_t}{t}
    \le2\sum_{t=1}^T u_t^2
    =2\mathbb E[R(T)].
\end{equation}
Let \(H_T:=\sum_{t=1}^T(t+D)^{-1}\).
Dividing~\eqref{eq:root-regret-recurrence} by \(t+D\), summing, and applying Cauchy--Schwarz and~\eqref{eq:weighted-regret-sum} gives
\[
    cD H_T
    \le C\sqrt{D\,\mathbb E[R(T)]H_T}+2\mathbb E[R(T)].
\]
The inequality \(2ab\le a^2+b^2\) gives
\[
    C\sqrt{D\,\mathbb E[R(T)]H_T}
    \le\frac c2 DH_T+\frac{C^2}{2c}\mathbb E[R(T)].
\]
Substituting this bound and rearranging gives \((c/2)DH_T\le(2+C^2/(2c))\mathbb E[R(T)]\), hence \(\mathbb E[R(T)]\ge c'DH_T\) for a universal constant \(c'>0\).
Since \(D\ge1\),
\[
    H_T\ge\log\left(1+\frac{T}{D+1}\right)
    \ge\frac12\log\left(1+\frac{T}{D}\right).
\]
Thus the regret averaged over the prior is at least \(c d^{3/2}\log(1+T/d^{3/2})\).
For each policy, some parameter in the support of this fixed prior attains at least this regret, proving the minimax claim.
\end{proof}

\section{Algorithms for Finite Action Sets}
\label{sec:finite-upper}

For fixed finite action sets, we propose \SoftGElim and \QGElim within the experimental design and phased elimination framework of~\citet{liu2026quantum}.
In each phase, both algorithms estimate the reward means of all surviving actions using weighted sums of mean estimates for a small subset selected by the design.
\SoftGElim independently applies a quantum mean estimator with small bias and controlled tails (Lemma~\ref{lem:soft-qme}) to each action in this subset.
\QGElim chooses between the estimation procedure used by \SoftGElim and an alternative based on QMC, using whichever has the smaller query budget.

\subsection{Estimating Action Means}
\label{subsec:design-reconstruction}

We first construct a quantum estimator for the mean reward of a single action by adapting the phase and probability estimation methods of~\citet[Sections~6.2--6.4]{vanapeldoorn2023tomography}.
Appendix~\ref{app:soft-qme} gives the construction and proof.

\begin{lemma}[Quantum mean estimation, adapted from~{\citealp{vanapeldoorn2023tomography}}]
\label{lem:soft-qme}
Let \(\mathcal O\) be a reward oracle as defined in Section~\ref{sec:preliminaries}, with mean \(\mu\).
There exists a quantum algorithm \(\SoftQME(\mathcal O,M,\eta)\) such that, for any power of two \(M\ge32\) and \(\eta\in(0,1/8]\), the algorithm outputs an estimate \(\widehat\mu\in[-1/6,7/6]\) of \(\mu\) satisfying
\begin{equation}
\label{eq:soft-scalar-guarantee}
    |\mathbb E\widehat\mu-\mu|\le\eta,
    \qquad
    \Pr\left[|\widehat\mu-\mu|>\frac{12}{M}\right]\le\eta.
\end{equation}
It uses at most \(O(M\log(1/\eta))\) queries in total to \(\mathcal O\) and \(\mathcal O^\dagger\).
\end{lemma}

We next use a small-support approximate \(G\)-optimal design to estimate the reward means of all actions by querying only a small subset of them.
For a finite set \(\mathcal X\subseteq\mathbb R^d\), write \(r:=\dim(\operatorname{span}(\mathcal X))\).

\begin{lemma}[Small-support approximate \(G\)-optimal design {\citep{allen2021near,liu2026quantum}}]
\label{lem:small-support-design}
There are universal constants \(c_s,C_G\ge1\) such that, for every finite set \(\mathcal X\subseteq\mathbb R^d\) with \(r\ge1\), one can compute a subset \(S\subseteq\mathcal X\) and a probability distribution \(\pi\) with support \(S\) for which
\[
    |S|\le c_s r,
    \qquad
    V:=\sum_{a\in S}\pi(a)aa^\top,
    \qquad
    \operatorname{Range}(V)=\operatorname{span}(\mathcal X),
    \qquad
    \max_{x\in\mathcal X}x^\top V^+x\le C_G r.
\]
For every \(x\in\mathcal X\), define \(\alpha_a(x):=\pi(a)a^\top V^+x\) for \(a\in S\), where \(V^+\) is the Moore--Penrose pseudoinverse of \(V\).
These coefficients satisfy
\begin{equation}
\label{eq:design-coefficients}
    x=\sum_{a\in S}\alpha_a(x)a,
    \qquad
    \sum_{a\in S}\frac{\alpha_a(x)^2}{\pi(a)}
    =x^\top V^+x\le C_G r.
\end{equation}
\end{lemma}

Since \(x=\sum_{a\in S}\alpha_a(x)a\) and \(\mu(x)=x^\top\theta\), we have \(\mu(x)=\sum_{a\in S}\alpha_a(x)\mu(a)\).
Fix a design \((S,\pi,V)\) satisfying Lemma~\ref{lem:small-support-design}, a target accuracy \(\epsilon\in(0,1]\), and a failure probability \(\delta\in(0,1/2]\).
We choose the estimation budgets for the support actions according to the design weights, allowing us to control the concentration error of the weighted estimates using~\eqref{eq:design-coefficients}.
We define the procedure \(\SoftDesignEstimate(\mathcal X,S,\pi,V,\epsilon,\delta)\) as follows.
Let \(A_0=48\sqrt{C_G}\) and \(c_0=[64(c_s+\sqrt{C_G}+1)]^{-1}\).
For each \(a\in S\), choose \(M_a\) to be the smallest power of two satisfying
{
\begin{equation}
    M_a\ge\max\left\{
        32,\frac{A_0\sqrt{r\pi(a)\log(8|\mathcal X|/\delta)}}{\epsilon}
    \right\}.
    \label{eq:soft-design-resolution}
\end{equation}
}
Set \(\eta:=c_0\min\{\delta/r,\epsilon/\sqrt r,1/8\}\).
For each \(a\in S\), this procedure independently calls \(\SoftQME(\mathcal O_a,M_a,\eta)\) to obtain \(\widehat\mu_a\).
It then returns \(\widehat\mu(x):=\sum_{a\in S}\alpha_a(x)\widehat\mu_a\) for every \(x\in\mathcal X\).

\begin{lemma}
\label{lem:soft-design}
The procedure \(\SoftDesignEstimate(\mathcal X,S,\pi,V,\epsilon,\delta)\) returns estimates satisfying \(\Pr[\max_{x\in\mathcal X}|\widehat\mu(x)-\mu(x)|>\epsilon]\le\delta\) and uses at most
\begin{equation}
\label{eq:soft-design-cost}
    C\left(r+\frac{r\sqrt{\log(8|\mathcal X|/\delta)}}{\epsilon}\right)
    \log\left(C\max\left\{\frac r\delta,
        \frac{\sqrt r}{\epsilon},2\right\}\right)
\end{equation}
queries to the reward oracles and their inverses, for a universal constant \(C>1\).
\end{lemma}

To bound \(\max_{x\in\mathcal X}|\widehat\mu(x)-\mu(x)|\), we condition on \(|\widehat\mu_a-\mu(a)|\le12/M_a\) for every \(a\in S\).
The bounded output range and tail guarantee ensure that conditioning on each estimate's accuracy event changes its mean by at most \(O(\eta)\).
The guarantees in Lemma~\ref{lem:soft-qme}, the coefficient bound~\eqref{eq:design-coefficients}, and the choice of \(\eta\) ensure that the conditional expectation of each \(\widehat\mu(x)\) differs from \(\mu(x)\) by at most \(\epsilon/4\).
Under this conditioning, the estimates \(\widehat\mu_a\) remain independent, and each lies in an interval of length \(24/M_a\).
With the chosen \(M_a\), Hoeffding's inequality and a union bound over \(\mathcal X\) control the deviations of all \(\widehat\mu(x)\) from these expectations.
Accounting also for the probability that the conditioning event fails gives simultaneous error at most \(\epsilon\) with probability at least \(1-\delta\).
Appendix~\ref{app:soft-design-proof} gives the full proof.

\subsection{Algorithms}
\label{subsec:finite-elimination}

The algorithm \SoftGElim operates in phases indexed by \(\ell=0,1,\ldots,\lceil\log_2T\rceil\), with target accuracy \(\epsilon_\ell=2^{-\ell}\).
In phase \(\ell\), it constructs a design for the surviving actions (Line~\ref{line:finite-design}), calls \(\SoftDesignEstimate\) to estimate their reward means (Line~\ref{line:finite-estimate}), and eliminates actions with estimates more than \(2\epsilon_\ell\) below the maximum (Line~\ref{line:finite-eliminate}).
The algorithm stops after \(T\) rounds, even if the current estimation call has not finished.

\begin{algorithm}[H]
\caption{\SoftGElim}
\label{alg:finite-elimination}
\KwIn{Finite action set \(\mathcal A\), horizon \(T\ge2\), confidence parameter \(\delta\in(0,1)\)}
\(L_T\leftarrow\lceil\log_2T\rceil\), \(\mathcal A_0\leftarrow\mathcal A\),
    \(\delta'\leftarrow\delta/(L_T+1)\)\;
\For{\(\ell=0,\ldots,L_T\)}{
    \(\epsilon_\ell\leftarrow2^{-\ell}\)\;
    \If{\(|\mathcal A_\ell|=1\)}{
        Select the remaining action until time \(T\) and stop\;
    }
    Compute a design \((S_\ell,\pi_\ell,V_\ell)\) for \(\mathcal A_\ell\) using Lemma~\ref{lem:small-support-design}\label{line:finite-design}\;
    \(\{\widehat\mu_\ell(x)\}_{x\in\mathcal A_\ell}
        \leftarrow\SoftDesignEstimate
        (\mathcal A_\ell,S_\ell,\pi_\ell,V_\ell,\epsilon_\ell,\delta')\)
        \label{line:finite-estimate}\;
    \(\widehat\mu_\ell^\star\leftarrow
        \max_{x\in\mathcal A_\ell}\widehat\mu_\ell(x)\)\;
    \(\mathcal A_{\ell+1}\leftarrow
        \{x\in\mathcal A_\ell:
        \widehat\mu_\ell(x)\ge\widehat\mu_\ell^\star-2\epsilon_\ell\}\)\label{line:finite-eliminate}\;
}
Select any action in \(\mathcal A_{L_T+1}\) for all remaining rounds\;
\end{algorithm}

\begin{theorem}
\label{thm:soft-elimination}
For every finite action set \(\mathcal A\) of size \(K\ge2\), every horizon \(T\ge2\), and every \(\delta\in(0,1)\), \SoftGElim (Algorithm~\ref{alg:finite-elimination}) satisfies, with probability at least \(1-\delta\),
{
\[
    R(T)
    =O\left(d\log T\sqrt{\log(KT/\delta)}\log(dT/\delta)\right).
\]
}
In particular, setting \(\delta=1/T\) gives \(\mathbb E[R(T)]=O(d\log T\sqrt{\log(KT)}\log(dT))\).
\end{theorem}

\begin{proof}
The algorithm executes at most \(L_T+1=\lceil\log_2T\rceil+1\) phases.
Let \(\mathcal E\) be the event that \(|\widehat\mu_\ell(x)-\mu(x)|\le\epsilon_\ell\) for every \(x\in\mathcal A_\ell\) in every completed phase.
For each estimation call in phase \(\ell\), let \(\mathcal H_\ell\) denote the history before this call.
Conditional on \(\mathcal H_\ell\), the surviving action set \(\mathcal A_\ell\) and the design \((S_\ell,\pi_\ell,V_\ell)\) are fixed, so Lemma~\ref{lem:soft-design} bounds the probability that the phase completes with an estimation error exceeding \(\epsilon_\ell\) by \(\delta'\).
Taking expectations and applying a union bound over at most \(L_T+1\) phases gives \(\Pr(\mathcal E^c)\le(L_T+1)\delta'=\delta\).

On \(\mathcal E\), whenever \(x^\star\in\mathcal A_\ell\), we have \(\widehat\mu_\ell^\star-\widehat\mu_\ell(x^\star)\le2\epsilon_\ell\), so induction shows that \(x^\star\) survives every completed phase.
For every \(x\in\mathcal A_{\ell+1}\), the elimination rule gives \(\Delta(x)\le\widehat\mu_\ell(x^\star)-\widehat\mu_\ell(x)+2\epsilon_\ell\le4\epsilon_\ell\)~\citep[proof of Theorem~5.4]{liu2026quantum}.
Each query in phase \(\ell\) therefore incurs regret at most \(8\epsilon_\ell\), since actions in \(\mathcal A_\ell\) have gap at most \(4\epsilon_{\ell-1}=8\epsilon_\ell\) for \(\ell\ge1\) and at most \(1=\epsilon_0\) for \(\ell=0\).

Let \(Q_\ell\) be the number of estimation queries actually made in phase \(\ell\), and let \(r_\ell:=\dim(\operatorname{span}(\mathcal A_\ell))\).
Since \(r_\ell\le d\), \(\epsilon_\ell^{-1}\le2T\), and \((\delta')^{-1}=(L_T+1)/\delta\le T/\delta\), we have
\[
    \log\left(C\max\left\{
        \frac{r_\ell}{\delta'},
        \frac{\sqrt{r_\ell}}{\epsilon_\ell},2
    \right\}\right)
    =O\left(\log(dT/\delta)\right).
\]
Using these bounds and \(|\mathcal A_\ell|\le K\) in~\eqref{eq:soft-design-cost} gives
\[
    Q_\ell\le
    C\left(d+\frac{d\sqrt{\log(KT/\delta)}}{\epsilon_\ell}\right)
    \log(dT/\delta).
\]
The regret incurred by these queries on \(\mathcal E\) is therefore at most
\[
    8\epsilon_\ell Q_\ell
    \le C d\left(\epsilon_\ell+\sqrt{\log(KT/\delta)}\right)
    \log(dT/\delta).
\]

Summing over at most \(L_T+1=O(\log T)\) phases and using \(\sum_{\ell=0}^{L_T}\epsilon_\ell\le2\) gives the stated regret bound for estimation queries on \(\mathcal E\).
The remaining rounds contribute zero regret if only one action survives and at most \(4T\epsilon_{L_T}\le4\) if all phases are completed.
Since \(\Pr(\mathcal E)\ge1-\delta\), the high probability bound follows.

Setting \(\delta=1/T\) gives the expected regret bound, since \(R(T)\le T\) and \(T\Pr(\mathcal E^c)\le1\).
\end{proof}

\paragraph{Choosing between estimation procedures.}
The algorithm \QGElim combines \SoftDesignEstimate with the QMC estimation procedure \QMCDesignEstimate of~\citet[Lemma~5.6]{liu2026quantum} to obtain a better regret bound.
\QMCDesignEstimate has a query bound independent of the number of surviving actions, while \SoftDesignEstimate has a better polynomial dependence on the dimension.
In each phase, \QGElim chooses the procedure with the smaller query budget to estimate reward means.

Fix \(\mathcal X\), a design \((S,\pi,V)\) for \(\mathcal X\), and target parameters \(\epsilon,\delta\) as in Section~\ref{subsec:design-reconstruction}.
Let \(Q_{\rm soft}\) and \(Q_{\rm QMC}\) denote the query budgets for the two procedures, respectively, which are computed as follows:
\begin{itemize}
    \item \SoftDesignEstimate. Lemma~\ref{lem:soft-design} gives the query budget
    \begin{equation}
    \begin{aligned}
        \label{eq:combined-query-budgets-soft}
        Q_{\rm soft}
        :=C\left(
            r+\frac{r\sqrt{\log(8|\mathcal X|/\delta)}}{\epsilon}
        \right)
        \log\left(C\max\left\{
            \frac r\delta,\frac{\sqrt r}{\epsilon},2
        \right\}\right).
    \end{aligned}
    \end{equation}
    \item \QMCDesignEstimate:
        For each \(a\in S\), call \(\QMC(\mathcal O_a,\rho,\delta/|S|)\) with \(\rho:=\epsilon/\sqrt{C_G r}\), and combine the outputs using the coefficients \(\alpha_a(x)\).
    By~\eqref{eq:design-coefficients} and Cauchy--Schwarz, \(\sum_{a\in S}|\alpha_a(x)|\le\sqrt{C_Gr}\) for every \(x\in\mathcal X\), so support estimates with error at most \(\rho\) give reconstruction error at most \(\rho\sqrt{C_Gr}=\epsilon\).
    A union bound over the support actions therefore ensures that all reward means in \(\mathcal X\) are estimated to within \(\epsilon\) with probability at least \(1-\delta\).
    Lemma~\ref{lem:qmc} gives the query budget
    \begin{equation}
    \begin{aligned}
        \label{eq:combined-query-budgets-qmc}
        Q_{\rm QMC}
        :=|S|\left\lceil\frac{C_{\rm QMC}}{\rho}
            \log\frac{|S|}{\delta}\right\rceil.
    \end{aligned}
    \end{equation}
\end{itemize}

We fix sufficiently large universal constants in these bounds so that both query budgets are valid deterministic upper bounds for their respective estimation procedures.
The algorithm \QGElim follows Algorithm~\ref{alg:finite-elimination}, replacing the estimation call in Line~\ref{line:finite-estimate} with the selected procedure.
Algorithm~\ref{alg:combined-elimination} in Appendix~\ref{app:estimation-budgets} gives the full procedure.

\begin{theorem}
\label{thm:combined-elimination}
For every finite action set of size \(K\ge2\), every \(T\ge2\), and every \(\delta\in(0,1)\), \QGElim satisfies, with probability at least \(1-\delta\),
{
\[
    R(T)
    =O\left(
        d\log T\log(dT/\delta)
        \min\left\{\sqrt{\log(KT/\delta)},\sqrt d\right\}
    \right).
\]
}
In particular, setting \(\delta=1/T\) gives \(\mathbb E[R(T)]=O(d\log T\log(dT)\min\{\sqrt{\log(KT)},\sqrt d\})\).
\end{theorem}

\begin{proof}
Define \(\mathcal E\) as in the proof of Theorem~\ref{thm:soft-elimination}.
Conditional on the history before each estimation call, the design and the selected procedure are fixed, and either procedure has failure probability at most \(\delta'\).
A union bound over at most \(L_T+1\) phases therefore gives \(\Pr(\mathcal E)\ge1-\delta\).

Following the proof of Theorem~\ref{thm:soft-elimination}, in phase \(\ell\), we have
\[
\begin{aligned}
    Q_{\rm soft}
    &\le C'\frac{d}{\epsilon_\ell}
        \sqrt{\log(KT/\delta)}\log(dT/\delta),\\
    Q_{\rm QMC}
    &\le C'\frac{d^{3/2}}{\epsilon_\ell}\log(dT/\delta),
\end{aligned}
\]
for a universal constant \(C'>0\).

On \(\mathcal E\), every action queried in phase \(\ell\) has gap at most \(8\epsilon_\ell\).
Since \QGElim uses the procedure with the smaller query budget, the regret incurred by these queries is at most
\[
    8\epsilon_\ell\min\{Q_{\rm soft},Q_{\rm QMC}\}
    \le 8C'd\log(dT/\delta)
        \min\left\{\sqrt{\log(KT/\delta)},\sqrt d\right\}.
\]

Summing over at most \(L_T+1=O(\log T)\) phases and accounting for the remaining rounds as in the proof of Theorem~\ref{thm:soft-elimination} proves the high probability bound.

Setting \(\delta=1/T\) gives the expected regret bound, since the failure event contributes at most \(T\delta=1\).
\end{proof}

\section{Conclusion}
We proved a minimax regret lower bound of \(\Omega(d^{3/2}\log(1+T/d^{3/2}))\) for quantum linear bandits with general action sets.
This shows that the \(d^{3/2}\) dependence in the best known upper bound is optimal up to logarithmic factors.
Our proof bounds posterior Fisher information in terms of the regret incurred, then applies a matrix Cram\'er--Rao argument to obtain a lower bound on the error in estimating the optimal direction and hence on regret.
For fixed finite action sets, \SoftGElim uses independent quantum mean estimates with small bias and controlled tails to reduce the \(\log(KT)\log\log(dT)\) factor in the previous regret bound to \(\sqrt{\log(KT)}\).
\QGElim further reduces this factor to \(\min\{\sqrt{\log(KT)},\sqrt d\}\) by choosing the estimation procedure with the smaller query budget in each phase.

These results suggest several directions for future work.
For general action sets, an open question is whether the remaining logarithmic gap between the upper and lower bounds can be closed.
For fixed finite action sets, lower bounds that capture the dependence on \(K\) would clarify the tightness of our upper bound, since the existing lower-bound construction uses only \(d\) actions~\citep{liu2026quantum}.
Our posterior Fisher information bound (Lemma~\ref{lem:posterior-fisher}) may also help establish query lower bounds for quantum estimation problems with adaptive access to oracles that encode affine functions of an unknown parameter.
On the algorithmic side, a natural question is whether the small bias and controlled tails of \SoftQME{} (Lemma~\ref{lem:soft-qme}) can improve regret bounds in other bandit settings.

\section*{Acknowledgment}
The authors used generative AI tools as an interactive aid in developing and checking parts of the proofs, searching for and comparing related work, and improving the clarity of the manuscript.
The authors take responsibility for the final manuscript.

\bibliography{reference}

@inproceedings{vanapeldoorn2023tomography,
  title={Quantum tomography using state-preparation unitaries},
  author={van Apeldoorn, Joran and Cornelissen, Arjan and Gily{\'e}n, Andr{\'a}s and Nannicini, Giacomo},
  booktitle={Proceedings of the 2023 Annual ACM-SIAM Symposium on Discrete Algorithms},
  pages={1265--1318},
  year={2023},
  doi={10.1137/1.9781611977554.ch47}
}

@article{liu2026quantum,
  title={Quantum Multi-Armed Bandits and Linear Bandits: Lower Bounds and Algorithms},
  author={Liu, Maoli and Li, Zhuohua and Lui, John C. S.},
  journal={arXiv preprint arXiv:2608.14319},
  year={2026}
}

@incollection{brassard2002quantum,
  title={Quantum amplitude amplification and estimation},
  author={Brassard, Gilles and H{\o}yer, Peter and Mosca, Michele and Tapp, Alain},
  booktitle={Quantum Computation and Quantum Information},
  series={Contemporary Mathematics},
  volume={305},
  pages={53--74},
  publisher={American Mathematical Society},
  year={2002}
}

@inproceedings{wang2021quantum,
  title={Quantum exploration algorithms for multi-armed bandits},
  author={Wang, Daochen and You, Xuchen and Li, Tongyang and Childs, Andrew M},
  booktitle={Proceedings of the AAAI Conference on Artificial Intelligence},
  volume={35},
  number={11},
  pages={10102--10110},
  year={2021}
}

@inproceedings{wan2023quantum,
  title     = {Quantum multi-armed bandits and stochastic linear bandits enjoy logarithmic regrets},
  author    = {Wan, Zongqi and Zhang, Zhijie and Li, Tongyang and Zhang, Jialin and Sun, Xiaoming},
  booktitle = {Proceedings of the AAAI Conference on Artificial Intelligence},
  volume    = {37},
  number    = {8},
  pages     = {10087--10094},
  year      = {2023}
}

@article{kiefer1960equivalence,
  title={The equivalence of two extremum problems},
  author={Kiefer, Jack and Wolfowitz, Jacob},
  journal={Canadian Journal of Mathematics},
  volume={12},
  pages={363--366},
  year={1960},
  publisher={Cambridge University Press}
}

@inproceedings{wang2025best,
  title = {Quantum Best Arm Identification with Quantum Oracles},
  author = {Wang, Xuchuang and Chen, Yu-Zhen Janice and Guedes de Andrade, Matheus and Allcock, Jonathan and Hajiesmaili, Mohammad and Lui, John C.S. and Towsley, Don},
  booktitle = {Proceedings of the AAAI Conference on Artificial Intelligence},
  volume = {39},
  number = {20},
  pages = {21321--21329},
  year = {2025},
  doi = {10.1609/aaai.v39i20.35432}
}

@article{montanaro2015quantum,
  title={Quantum speedup of Monte Carlo methods},
  author={Montanaro, Ashley},
  journal={Proceedings of the Royal Society A: Mathematical, Physical and Engineering Sciences},
  volume={471},
  number={2181},
  pages={20150301},
  year={2015},
  publisher={The Royal Society}
}

@article{chiribella2009realization,
  title={Realization schemes for quantum instruments in finite dimensions},
  author={Chiribella, Giulio and D'Ariano, Giacomo Mauro and Perinotti, Paolo},
  journal={Journal of Mathematical Physics},
  volume={50},
  number={4},
  pages={042101},
  year={2009},
  doi={10.1063/1.3105923}
}

@book{kallenberg2017random,
  title={Random Measures, Theory and Applications},
  author={Kallenberg, Olav},
  series={Probability Theory and Stochastic Modelling},
  volume={77},
  publisher={Springer},
  year={2017},
  doi={10.1007/978-3-319-41598-7}
}

@article{dembo1991information,
  title={Information theoretic inequalities},
  author={Dembo, Amir and Cover, Thomas M. and Thomas, Joy A.},
  journal={IEEE Transactions on Information Theory},
  volume={37},
  number={6},
  pages={1501--1518},
  year={1991},
  doi={10.1109/18.104312}
}

@inproceedings{cornelissen2023sublinear,
  title={A sublinear-time quantum algorithm for approximating partition functions},
  author={Cornelissen, Arjan and Hamoudi, Yassine},
  booktitle={Proceedings of the 2023 annual ACM-Siam symposium on discrete algorithms (SODA)},
  pages={1245--1264},
  year={2023},
  organization={SIAM}
}

@inproceedings{hamoudi2021quantum,
  title={{Quantum Sub-Gaussian Mean Estimator}},
  author={Hamoudi, Yassine},
  booktitle={29th Annual European Symposium on Algorithms (ESA 2021)},
  series={Leibniz International Proceedings in Informatics (LIPIcs)},
  volume={204},
  pages={50:1--50:17},
  year={2021},
  doi={10.4230/LIPIcs.ESA.2021.50}
}

@article{demkowicz2017adaptive,
  title = {Adaptive Quantum Metrology under General Markovian Noise},
  author = {Demkowicz-Dobrza{\ifmmode \acute{n}\else \'{n}\fi{}}ski, Rafa{\l{}} and Czajkowski, Jan and Sekatski, Pavel},
  journal = {Phys. Rev. X},
  volume = {7},
  issue = {4},
  pages = {041009},
  numpages = {15},
  year = {2017},
  month = {Oct}
}

@inproceedings{cornelissen2022near,
  title={Near-optimal quantum algorithms for multivariate mean estimation},
  author={Cornelissen, Arjan and Hamoudi, Yassine and Jerbi, Sofiene},
  booktitle={Proceedings of the 54th Annual ACM SIGACT Symposium on Theory of Computing},
  pages={33--43},
  year={2022}
}

@article{wu2023quantum,
  title={Quantum heavy-tailed bandits},
  author={Wu, Yulian and Guan, Chaowen and Aggarwal, Vaneet and Wang, Di},
  journal={arXiv preprint arXiv:2301.09680},
  year={2023}
}

@article{dai2023quantum,
  title={Quantum bayesian optimization},
  author={Dai, Zhongxiang and Lau, Gregory Kang Ruey and Verma, Arun and Shu, Yao and Low, Bryan Kian Hsiang and Jaillet, Patrick},
  journal={Advances in neural information processing systems},
  volume={36},
  pages={20179--20207},
  year={2023}
}

@inproceedings{hikima2024quantum,
  title={Quantum kernelized bandits},
  author={Hikima, Yasunari and Murao, Kazunori and Takemori, Sho and Umeda, Yuhei},
  booktitle={The 40th Conference on Uncertainty in Artificial Intelligence},
  year={2024}
}

@inproceedings{yi2026quantum,
  title={Quantum lipschitz bandits},
  author={Yi, Bongsoo and Kang, Yue and Li, Yao},
  booktitle={Proceedings of the AAAI Conference on Artificial Intelligence},
  volume={40},
  number={33},
  pages={27844--27851},
  year={2026}
}

@article{allen2021near,
  title={Near-optimal discrete optimization for experimental design: A regret minimization approach},
  author={Allen-Zhu, Zeyuan and Li, Yuanzhi and Singh, Aarti and Wang, Yining},
  journal={Mathematical Programming},
  volume={186},
  number={1},
  pages={439--478},
  year={2021},
  publisher={Springer}
}

@book{lattimore2020bandit,
  title={Bandit algorithms},
  author={Lattimore, Tor and Szepesv{\'a}ri, Csaba},
  year={2020},
  publisher={Cambridge University Press}
}

@inproceedings{lattimore2020learning,
  title={Learning with good feature representations in bandits and in rl with a generative model},
  author={Lattimore, Tor and Szepesvari, Csaba and Weisz, Gellert},
  booktitle={International conference on machine learning},
  pages={5662--5670},
  year={2020},
  organization={PMLR}
}

@article{bubeck2012regret,
  title={Regret analysis of stochastic and nonstochastic multi-armed bandit problems},
  author={Bubeck, S{\'e}bastien and Cesa-Bianchi, Nicol{\`o}},
  journal={Foundations and Trends{\textregistered} in Machine Learning},
  volume={5},
  number={1},
  pages={1--122},
  year={2012}
}

@inproceedings{awerbuch2004adaptive,
  title={Adaptive routing with end-to-end feedback: Distributed learning and geometric approaches},
  author={Awerbuch, Baruch and Kleinberg, Robert D},
  booktitle={Proceedings of the thirty-sixth annual ACM symposium on Theory of computing},
  pages={45--53},
  year={2004}
}

@inproceedings{kothari2023mean,
  title={Mean estimation when you have the source code; or, quantum Monte Carlo methods},
  author={Kothari, Robin and O'Donnell, Ryan},
  booktitle={Proceedings of the 2023 Annual ACM-SIAM Symposium on Discrete Algorithms (SODA)},
  pages={1186--1215},
  year={2023},
  organization={SIAM}
}

@article{jin2026nearly,
  title={Nearly Minimax-Optimal Regret for Linear Contextual Bandits with Arbitrary Adaptive Action Sets},
  author={Jin, Tianyuan},
  journal={arXiv preprint arXiv:2609.15170},
  year={2026}
}

@inproceedings{cheung2019learning,
  title={Learning to optimize under non-stationarity},
  author={Cheung, Wang Chi and Simchi-Levi, David and Zhu, Ruihao},
  booktitle={The 22nd International Conference on Artificial Intelligence and Statistics},
  pages={1079--1087},
  year={2019},
  organization={PMLR}
}

@inproceedings{li2010contextual,
  title={A contextual-bandit approach to personalized news article recommendation},
  author={Li, Lihong and Chu, Wei and Langford, John and Schapire, Robert E},
  booktitle={Proceedings of the 19th international conference on World wide web},
  pages={661--670},
  year={2010}
}

@article{abbasi2011improved,
  title={Improved algorithms for linear stochastic bandits},
  author={Abbasi-Yadkori, Yasin and P{\'a}l, D{\'a}vid and Szepesv{\'a}ri, Csaba},
  journal={Advances in neural information processing systems},
  volume={24},
  year={2011}
}

@inproceedings{dani2008stochastic,
  title={Stochastic linear optimization under bandit feedback},
  author={Dani, Varsha and Hayes, Thomas P and Kakade, Sham M},
  booktitle={21st Annual Conference on Learning Theory},
  number={101},
  pages={355--366},
  year={2008}
}

@inproceedings{li2019nearly,
  title={Nearly minimax-optimal regret for linearly parameterized bandits},
  author={Li, Yingkai and Wang, Yining and Zhou, Yuan},
  booktitle={Conference on Learning Theory},
  pages={2173--2174},
  year={2019},
  organization={PMLR}
}

@article{buchholz2025multi,
  title={Multi-Armed Bandits and Quantum Channel Oracles},
  author={Buchholz, Simon and K{\"u}bler, Jonas M and Sch{\"o}lkopf, Bernhard},
  journal={Quantum},
  volume={9},
  pages={1672},
  year={2025},
  publisher={Verein zur F{\"o}rderung des Open Access Publizierens in den Quantenwissenschaften}
}

@article{su2025quantum,
  title={Quantum Algorithms for Bandits with Knapsacks with Improved Regret and Time Complexities},
  author={Su, Yuexin and Yang, Ziyi and Huang, Peiyuan and Li, Tongyang and Ye, Yinyu},
  journal={arXiv preprint arXiv:2507.04438},
  year={2025}
}

@article{lumbreras2022multi,
  title={Multi-armed quantum bandits: Exploration versus exploitation when learning properties of quantum states},
  author={Lumbreras, Josep and Haapasalo, Erkka and Tomamichel, Marco},
  journal={Quantum},
  volume={6},
  pages={749},
  year={2022},
  publisher={Verein zur F{\"o}rderung des Open Access Publizierens in den Quantenwissenschaften}
}

@inproceedings{lumbreras2024linear,
  title={Linear bandits with polylogarithmic minimax regret},
  author={Lumbreras, Josep and Tomamichel, Marco},
  booktitle={The Thirty Seventh Annual Conference on Learning Theory},
  pages={3644--3682},
  year={2024},
  organization={PMLR}
}

@inproceedings{chu2011contextual,
  title={Contextual bandits with linear payoff functions},
  author={Chu, Wei and Li, Lihong and Reyzin, Lev and Schapire, Robert},
  booktitle={Proceedings of the fourteenth international conference on artificial intelligence and statistics},
  pages={208--214},
  year={2011},
  organization={JMLR Workshop and Conference Proceedings}
}

@article{lumbreras2026learning,
  title={Learning pure quantum states almost without regret},
  author={Lumbreras, Josep and Terekhov, Mikhail and Tomamichel, Marco},
  journal={npj Quantum Information},
  year={2026},
  publisher={Nature Publishing Group UK London}
}

@article{jordan2005fast,
  title={Fast quantum algorithm for numerical gradient estimation},
  author={Jordan, Stephen P.},
  journal={Physical Review Letters},
  volume={95},
  number={5},
  pages={050501},
  year={2005},
  doi={10.1103/PhysRevLett.95.050501}
}

@inproceedings{gilyen2019optimizing,
  title={Optimizing quantum optimization algorithms via faster quantum gradient computation},
  author={Gily{\'e}n, Andr{\'a}s and Arunachalam, Srinivasan and Wiebe, Nathan},
  booktitle={Proceedings of the Thirtieth Annual ACM-SIAM Symposium on Discrete Algorithms},
  pages={1425--1444},
  year={2019},
  doi={10.1137/1.9781611975482.87}
}

@inproceedings{li2022quantum,
  title={Quantum speedups of optimizing approximately convex functions with applications to logarithmic regret stochastic convex bandits},
  author={Li, Tongyang and Zhang, Ruizhe},
  booktitle={Advances in Neural Information Processing Systems},
  volume={35},
  pages={3152--3164},
  year={2022}
}
\bibliographystyle{abbrvnat}

\clearpage
\appendix
\section{Proofs for Section~\ref{sec:general-lower}}
\label{app:posterior-fisher}

\subsection{Proof of Lemma~\ref{lem:posterior-fisher}}
\label{app:posterior-fisher-proof}

Throughout the paper, all measurement outcome spaces are assumed to be standard Borel.
For each policy or protocol, the quantum memory lies in a fixed finite-dimensional Hilbert space throughout its execution.
The dimension may depend on the policy or protocol and the problem parameters.

\begin{proof}
We first give a brief overview of the proof.
We use quantum instruments to represent the operations between successive oracle queries, capturing both classical measurement outcomes and changes to quantum memory.
We then refine the classical record so that, for each fixed parameter value \(z\), every complete refined record \(y\) determines an unnormalized pure state \(f_y(z)\).
The squared norm of this state gives the conditional probability density of \(y\) given \(Z=z\).
Next, we differentiate \(f_y(z)\) in the directions spanned by the columns of \(B\) and bound these derivatives by separating the contributions of individual oracle queries.
Finally, we combine the contributions of the record and the prior to bound the posterior value of \(S_B\).

\textbf{Step 1. Representing the posterior density.}
To obtain the posterior density by Bayes' rule, we first construct a refined classical record and express its conditional density given the parameter.

We group the operations before, between, and after oracle queries, splitting these groups at round boundaries when needed.
For the classical record \(h\) preceding a group, we represent its operations by a quantum instrument \(\mathcal I_h\) acting on the quantum memory.
The classical outcome \(u\) records all measurement outcomes and the policy's random choices during these operations.
If the protocol terminates at this stage, \(u\) also includes any final classical output.

To construct the refined record, we express each quantum instrument in terms of Kraus operators.
Let \(n\) be the dimension of the quantum memory, and fix a basis.
For each measurable outcome set \(E\), let \(C_h(E)\) be the Choi matrix of \(\mathcal I_h(E)\) formed using \(\sum_{i=1}^n|i\rangle|i\rangle\).
Following~\citet[Proposition~2]{chiribella2009realization}, we choose the scalar reference measure \(\nu_h(E):=\operatorname{tr}C_h(E)\).
The instruments' measurable dependence on \(h\) makes \(\nu_h\) a kernel, and trace preservation gives it total mass \(n\).
For each fixed \(\xi\in\mathbb C^{n^2}\), the nonnegative scalar kernel \(E\mapsto\xi^\dagger C_h(E)\xi\) is dominated by \(\|\xi\|_2^2\nu_h\).

Since the outcome space is standard Borel, the Radon--Nikodym theorem for finite kernels~\citep[Theorem~1.28]{kallenberg2017random} applies to these scalar kernels.
Applying it to the finitely many kernels needed for polarization yields a jointly measurable matrix density \(C_{h,u}\) satisfying \(C_h(E)=\int_E C_{h,u}\,\nu_h(du)\).
For each \(h\), this density is finite and positive semidefinite for \(\nu_h\)-almost every \(u\), and we set it to zero on the measurable exceptional set.
Taking the positive semidefinite square root of \(C_{h,u}\) and reshaping its columns gives Kraus operators \(K_{h,u,j}\), \(j=1,\ldots,n^2\).
These operators are jointly measurable in \(h,u\) because the matrix square root is continuous on positive semidefinite matrices.
For an input density operator \(\rho\) and a measurable outcome set \(E\), these operators give the unnormalized output state \(\mathcal I_h(E)(\rho)\) and satisfy the completeness relation
\[
    \begin{aligned}
        &\mathcal I_h(E)(\rho)=\int_E\sum_j K_{h,u,j}\rho K_{h,u,j}^{\dagger}\,\nu_h(du),\\
        &\int\sum_j K_{h,u,j}^{\dagger}K_{h,u,j}\,\nu_h(du)=I,
    \end{aligned}
\]
where \(I\) is the identity operator on the quantum memory.
We refine each outcome from \(u\) to \((u,j)\), where \(j\) indexes the Kraus operators and is hidden from the policy.
Summing over \(j\) recovers the original outcome distribution and quantum evolution.
For each history, \(\nu_h\) together with counting measure on the finite Kraus index set defines a finite reference measure on the refined outcomes \((u,j)\).
We normalize this measure and rescale the Kraus operators accordingly, preserving the instrument.
We construct a probability measure \(\nu(dy)\) on complete refined records \(y\) sequentially, using at each stage the normalized reference measure determined by the preceding record.
The refinement and reference measure depend only on the protocol and are independent of the unknown parameter, the prior, and \(B\).

To describe the quantum evolution using state vectors, we first purify the initial parameter-independent state.
We do so by adding a register on which all subsequent operations act as the identity.
Let \(\mathcal H\) denote the joint Hilbert space of the quantum memory and this added register.
For each fixed parameter value \(z\) and complete refined record \(y\), applying the corresponding oracle gates and Kraus operators to the purified initial state gives an unnormalized vector \(f_y(z)\in\mathcal H\).
For a round with no query, we use an identity gate in this product.
The conditional density of the refined record \(Y\) given \(Z=z\), relative to \(\nu\), is \(\ell_y(z):=\|f_y(z)\|^2\).
Under the prior \(p\), integration and Bayes' rule give the marginal density \(P_y\) of \(Y\) and the posterior density \(p_y\) of \(Z\) given \(Y=y\),
\[
    P_y:=\int p(z)\ell_y(z)\,dz,
    \qquad
    p_y(z)=\frac{p(z)\ell_y(z)}{P_y}
    \quad(P_y>0).
\]
We have \(\int\ell_y(z)\,\nu(dy)=1\) for every \(z\in\operatorname{supp}(p)\), and the law of \(Y\) is \(P_y\nu(dy)\).
Records with \(P_y=0\) can be omitted.

For a fixed \(y\), the action choices and Kraus operators are independent of \(z\), so \(f_y\) is a finite product of fixed matrices and continuously differentiable oracle gates.
Consequently, \(\|f_y\|\) is bounded and Lipschitz near the compact support of \(p\).
Multiplying \(\|f_y\|\) by a smooth cutoff with compact support in this neighborhood and equal to one near \(\operatorname{supp}(p)\) gives a bounded Lipschitz function on \(\mathbb R^m\) without changing \(\sqrt p\,\|f_y\|\).
The Sobolev product rule then implies that \(\sqrt{p_y}=P_y^{-1/2}\sqrt p\,\|f_y\|\) has a square-integrable weak gradient.

We compare prior and posterior Fisher information through their scores.
For a density \(r\) whose square root has a square-integrable weak gradient, define its score by \(s_r=2\nabla\sqrt r/\sqrt r\) on \(\{r>0\}\) and zero elsewhere.
For a differentiable density, this agrees with the gradient \(\nabla\log r\) of the log density on \(\{r>0\}\).
The weak gradient of \(\sqrt r\) vanishes almost everywhere on its zero set, so \(J(r)=\int r\,s_rs_r^\top\).
Where the posterior density is positive, \(s_{p_y}=s_p+\nabla\log\ell_y\).

\textbf{Step 2. Bounding each query contribution.}
To bound the likelihood derivatives in the directions spanned by \(B\), we differentiate the oracle gates in the product defining \(f_y\).
We control each derivative by averaging over later operations, which preserves the trace, as in~\citet[Appendix~B.4, Eqs.~(B9)--(B10)]{demkowicz2017adaptive}.
Let \(y_{<s}\) denote the refined record before the possible oracle query in round \(s\), including the selected action and the choice of whether to make a query.
For a direction \(u\in\mathbb R^k\), the product rule gives
\begin{equation}
\label{eq:app-branch-derivative}
    \partial_{Bu} f_y(z)
    =\sum_{s=1}^q
        \bigl(a_{x_s(y_{<s})}^\top Bu\bigr)w_{y,s}(z).
\end{equation}
Here \(w_{y,s}\) is obtained by replacing the oracle gate in round \(s\) with its derivative with respect to the mean.
If no query is made in that round, set \(w_{y,s}=0\) and interpret \(a_{x_s}=0\).
The derivative of an inverse query has the same operator norm as the forward derivative.

We first bound the contribution of one query after averaging over the record.
Let \(h=y_{<s}\), and let \(\psi_h(z)\) be the corresponding unnormalized state.
Write \(D_h(z)\) for the derivative of the oracle gate in round \(s\), or zero if no query is made, and \(V_{y\mid h}(z)\) for the product of all subsequent operators specified by \(y\).
Then \(w_{y,s}(z)=V_{y\mid h}(z)D_h(z)\psi_h(z)\) and \(\|D_h(z)\|_{\mathrm{op}}\le L\).
For fixed \(z,h\), integrating out all subsequent outcomes gives a map that preserves the trace.
Write \(y\succeq h\) when \(y\) extends \(h\), and let \(\nu(dy\mid h)\) be the reference measure for the remaining outcomes.
Applying trace preservation to the unnormalized state represented by \(D_h(z)\psi_h(z)\) gives
\begin{equation}
\label{eq:app-suffix-completeness}
    \int_{y\succeq h}\|w_{y,s}(z)\|^2\,\nu(dy\mid h)
    =\|D_h(z)\psi_h(z)\|^2.
\end{equation}

To combine the query contributions, we use the nuclear norm \(\|Q\|_*:=\operatorname{tr}\sqrt{Q^*Q}\), where \(Q^*\) is the adjoint.
This norm equals \(S_B(r)\) when \(Q^*Q=B^\top J(r)B\), and its triangle inequality allows us to add the contributions.
Since the parameter directions are real, we treat the spaces of quantum states as real Hilbert spaces with inner product \(\operatorname{Re}\langle\cdot,\cdot\rangle\).
Define \(Q_{y,s}\) as an operator from \(\mathbb R^k\) to the space of \(\mathcal H\)-valued functions square-integrable with respect to \(dz\) by
\[
    (Q_{y,s}u)(z)
    :=\bigl(a_{x_s(y_{<s})}^\top Bu\bigr)\sqrt{p(z)}\,w_{y,s}(z).
\]
For fixed \(y\), the action \(x_s(y_{<s})\) is fixed, so \(Q_{y,s}\) has rank at most one and
\[
    \|Q_{y,s}\|_*^2
    =\|B^\top a_{x_s(y_{<s})}\|_2^2
        \int p(z)\|w_{y,s}(z)\|^2\,dz.
\]
Let \(\nu_{<s}(dh)\) be the marginal reference measure on these preceding records.
Integrating first over the subsequent outcomes using~\eqref{eq:app-suffix-completeness}, and then over \(h\), gives
\begin{align}
    \int\|Q_{y,s}\|_*^2\,\nu(dy)
    &=\int p(z)\int
        \|B^\top a_{x_s(h)}\|_2^2
        \|D_h(z)\psi_h(z)\|^2\,\nu_{<s}(dh)\,dz\notag\\
    &\le L^2\int p(z)\int
        \|B^\top a_{x_s(h)}\|_2^2
        \|\psi_h(z)\|^2\,\nu_{<s}(dh)\,dz\notag\\
    &=L^2\mathbb E\|B^\top a_{x_s}\|_2^2.
    \label{eq:app-query-derivative}
\end{align}
The last expectation is under the original protocol because summing out the additional Kraus labels preserves its action distribution.

\textbf{Step 3. Bounding the likelihood contribution.}
Set \(Q_y=\sum_{s=1}^qQ_{y,s}\), so that \((Q_yu)(z)=\sqrt{p(z)}\,\partial_{Bu}f_y(z)\).
For each \(y\), the triangle inequality for the nuclear norm gives \(\|Q_y\|_*\le\sum_{s=1}^q\|Q_{y,s}\|_*\).
Minkowski's inequality in \(L_2(\nu)\), together with~\eqref{eq:app-query-derivative}, then gives
\begin{equation}
\label{eq:app-total-derivative}
\begin{aligned}
    \left(\int\|Q_y\|_*^2\,\nu(dy)\right)^{1/2}
    &\le\sum_{s=1}^q
        \left(\int\|Q_{y,s}\|_*^2\,\nu(dy)\right)^{1/2}\\
    &\le L\sum_{s=1}^q
        \left(\mathbb E\|B^\top a_{x_s}\|_2^2\right)^{1/2}.
\end{aligned}
\end{equation}
We now use this bound to control the likelihood score in the same directions.
Cauchy--Schwarz gives
\[
    \partial_{Bu}\ell_y
    =2\operatorname{Re}\langle f_y,\partial_{Bu}f_y\rangle,
    \qquad
    \frac{(\partial_{Bu}\ell_y)^2}{\ell_y}
    \le4\|\partial_{Bu}f_y\|^2
\]
on \(\{\ell_y>0\}\).
We set the fraction and \(\nabla\log\ell_y\) to zero on \(\{\ell_y=0\}\).
Let \(L_2(p_y)\) be the space of real functions square-integrable with respect to \(p_y(z)\,dz\).
We represent the projected prior and likelihood scores by the operators \(A_y,C_y:\mathbb R^k\to L_2(p_y)\), respectively, defined by \((A_yu)(z):=u^\top B^\top s_p(z)\) and \((C_yu)(z):=u^\top B^\top\nabla\log\ell_y(z)\).
Then
\[
    C_y^*C_y
    =\frac1{P_y}\int p(z)
        \frac{B^\top\nabla\ell_y(z)\nabla\ell_y(z)^\top B}{\ell_y(z)}\,dz
    \preceq\frac4{P_y}Q_y^*Q_y.
\]
Monotonicity of the matrix square root and Cauchy--Schwarz therefore imply
\begin{equation}
\label{eq:app-likelihood-score}
    \int P_y\|C_y\|_*\,\nu(dy)
    \le2\int\sqrt{P_y}\|Q_y\|_*\,\nu(dy)
    \le2\left(\int\|Q_y\|_*^2\,\nu(dy)\right)^{1/2}.
\end{equation}

\textbf{Step 4. Combining prior and likelihood contributions.}
For the prior contribution, the identity \(\int\ell_y(z)\,\nu(dy)=1\) gives \(\int P_yA_y^*A_y\,\nu(dy)=B^\top J(p)B\).
Concavity of \(M\mapsto\operatorname{tr}\sqrt M\) yields
\begin{equation}
\label{eq:app-prior-score}
    \int P_y\|A_y\|_*\,\nu(dy)\le S_B(p).
\end{equation}
These bounds show that both operators are well defined for almost every record with \(P_y>0\).
The score decomposition \(s_{p_y}=s_p+\nabla\log\ell_y\) and the triangle inequality for the nuclear norm give
\[
    \begin{aligned}
        B^\top J(p_y)B&=(A_y+C_y)^*(A_y+C_y),\\
        S_B(p_y)&=\|A_y+C_y\|_*\le\|A_y\|_*+\|C_y\|_*.
    \end{aligned}
\]
Integrating with respect to the law \(P_y\nu(dy)\) of \(Y\) and applying~\eqref{eq:app-total-derivative}--\eqref{eq:app-prior-score} proves~\eqref{eq:posterior-fisher-bound}.
\end{proof}

\subsection{The Hard Instances}
\label{app:cap-instance}

We verify the properties of the action set, prior, and reward oracles used in the proof of Theorem~\ref{thm:general-lower}.
Every \(\theta\in\mathcal Q=\frac12e_1+[-w,w]^d\), with \(w=1/(8\sqrt d)\), satisfies \(\|\theta-\frac12e_1\|_2\le1/8\).
Consequently, \(3/8\le\|\theta\|_2\le5/8\), \(\theta_1\ge3/8\), and \(\|\theta_{2:d}\|_2\le1/8\).
The angle of \(g(\theta)=\theta/\|\theta\|_2\) with \(e_1\) is at most \(\arctan(1/3)<\pi/6\), so this action belongs to \(\mathcal A\).
Every \(x\in\mathcal A\) has \(x_1\ge\sqrt3/2\) and \(\|x_{2:d}\|_2\le1/2\).
Thus \((3\sqrt3-1)/16\le x^\top\theta\le\|\theta\|_2\le5/8\), which places all means in \([1/4,3/4]\).

The prior density is
\[
    p_0(\theta)
    =w^{-d}\prod_{j=1}^d
        \cos^2\left(
            \frac{\pi(\theta_j-\frac12\mathbf1\{j=1\})}{2w}
        \right)\mathbf1\{\theta\in\mathcal Q\}.
\]
Each cosine factor in \(\sqrt{p_0}\) vanishes at the endpoints of its interval.
Extending \(\sqrt{p_0}\) by zero outside \(\mathcal Q\) therefore gives a Lipschitz function with compact support, so it has a square-integrable weak gradient.
For the scalar density \(\varphi(z)=\cos^2(\pi z/2)\mathbf1\{|z|\le1\}\),
\[
    \int_{-1}^1\frac{\varphi'(z)^2}{\varphi(z)}\,dz
    =\pi^2\int_{-1}^1\sin^2(\pi z/2)\,dz
    =\pi^2.
\]
Scaling each coordinate by \(w\) gives \(\operatorname{tr}J(p_0)=d\pi^2/w^2=64\pi^2d^2\).

For a mean \(\mu\), define the one-qubit rotation \(V_\mu\) and the two-qubit reward unitary \(U_\mu\) by
\[
    V_\mu=
    \begin{pmatrix}
        \sqrt{1-\mu}&-\sqrt\mu\\
        \sqrt\mu&\sqrt{1-\mu}
    \end{pmatrix},
    \qquad
    U_\mu
    :=\mathrm{CNOT}_{2\to1}(I_2\otimes V_\mu).
\]
Set \(\mathcal O_x=U_{x^\top\theta}\).
The controlled-NOT gate copies the computational basis value of the reward qubit into the first register.
Hence \(\mathcal O_x|00\rangle=\sqrt{1-\mu(x)}\,|00\rangle+\sqrt{\mu(x)}\,|11\rangle\), as required by~\eqref{eq:reward-oracle}.
For \(\mu\in[1/4,3/4]\), direct differentiation gives \(\|\partial_\mu U_\mu\|_{\mathrm{op}}=\|\partial_\mu V_\mu\|_{\mathrm{op}}=1/(2\sqrt{\mu(1-\mu)})\le2/\sqrt3\).
The same derivative bound holds for the inverse oracle.
The bounds on the means place them strictly inside \((1/4,3/4)\), uniformly over \(x\in\mathcal A\) and \(\theta\in\mathcal Q\).
This interval therefore also contains all means in a sufficiently small neighborhood of \(\mathcal Q\), and the oracle family is continuously differentiable with the same derivative bound throughout the interval.

We next verify the properties of the reference direction used in the main proof.
Let \(\widetilde\theta\) be the Euclidean projection of \(W\) onto \(\mathcal Q\), so \(v=\widetilde\theta/\|\widetilde\theta\|_2\in\mathcal A\).
Projection gives \(\|\widetilde\theta-\theta\|_2\le\|W-\theta\|_2\), and the lower bound \(\|\theta\|_2,\|\widetilde\theta\|_2\ge3/8\) implies
\[
    \|v-g(\theta)\|_2
    \le\frac{2\|\widetilde\theta-\theta\|_2}
        {\min\{\|\widetilde\theta\|_2,\|\theta\|_2\}}
    \le\frac{16}{3}\|W-\theta\|_2.
\]
Because \(v_1>0\), the projections of \(e_2,\ldots,e_d\) onto \(v^\perp\) are linearly independent.
Applying Gram--Schmidt in this order gives an orthonormal basis matrix \(B(v)\) continuous on \(\mathcal A\), and hence measurable in \(W\).
Finally, the condition \(x_1\ge\sqrt3/2\) places every \(x\in\mathcal A\) within angle \(\pi/6\) of \(e_1\).
The angle between \(v\) and \(g(\theta)\) is therefore at most \(\pi/3\), giving \(v^\top g(\theta)\ge1/2\).

\subsection{Estimating the Optimal Direction}
\label{app:direction-estimation}

We prove~\eqref{eq:posterior-direction-error} by an integration-by-parts and matrix Cauchy--Schwarz argument for the nonlinear map \(g(\theta)=\theta/\|\theta\|_2\).
This is the form of the Cram\'er--Rao method needed here, with the derivative of \(g\) replacing the identity matrix in the covariance bound~\citep[Theorem~20]{dembo1991information}.
Fix \(v,x\in\mathcal A\), let \(B\) be an orthonormal basis matrix for \(v^\perp\), and let \(p\) be any probability density supported on \(\mathcal Q\) whose square root has a square-integrable weak gradient.
Write \(k=d-1\), and define
\[
    e(\theta):=B^\top(x-g(\theta)),
    \qquad
    s_B(\theta):=B^\top s_p(\theta),
    \qquad
    J_B:=B^\top J(p)B.
\]
The score \(s_p\) is defined as in Appendix~\ref{app:posterior-fisher-proof}, with value zero on \(\{p=0\}\), so \(J_B=\mathbb E_p[s_Bs_B^\top]\).

To justify integration by parts, note that \(\nabla p=2\sqrt p\,\nabla\sqrt p\) is integrable by Cauchy--Schwarz.
The map \(g\) is smooth near \(\mathcal Q\), since every point in this cube has norm at least \(3/8\).
Multiplying \(e\) by a smooth cutoff supported away from the origin and equal to one near \(\mathcal Q\) gives a compactly supported smooth test function without changing its values or derivatives on the support of \(p\).
Weak integration by parts therefore yields
\begin{equation}
\label{eq:app-direction-score}
    \mathbb E_p[e\,s_B^\top]
    =\int e(\theta)\nabla p(\theta)^\top B\,d\theta
    =\mathbb E_p[B^\top Dg(\theta)B]
    =:M.
\end{equation}
Since \(Dg(\theta)=(I_d-g(\theta)g(\theta)^\top)/\|\theta\|_2\) and \(\|B^\top g(\theta)\|_2^2=1-(v^\top g(\theta))^2\),
\[
    B^\top Dg(\theta)B
    \succeq\frac{(v^\top g(\theta))^2}{\|\theta\|_2}I_k
    \succeq\frac25 I_k.
\]
Thus \(M\succeq(2/5)I_k\).
The matrix \(J_B\) is positive definite.
Indeed, if \(u^\top J_Bu=0\), then \(s_B^\top u=0\) almost surely and~\eqref{eq:app-direction-score} gives \(Mu=0\), which forces \(u=0\).

Expanding the positive semidefinite matrix \(\mathbb E_p[(e-MJ_B^{-1}s_B)(e-MJ_B^{-1}s_B)^\top]\) gives
\[
    \mathbb E_p[ee^\top]\succeq M J_B^{-1}M^\top.
\]
Taking traces and using \(M^\top M\succeq(4/25)I_k\), we obtain
\[
    \mathbb E_p\|e\|_2^2
    \ge\operatorname{tr}(J_B^{-1}M^\top M)
    \ge\frac4{25}\operatorname{tr}(J_B^{-1}).
\]
If \(\gamma_1,\ldots,\gamma_k>0\) are the eigenvalues of \(J_B\), H\"older's inequality gives
\[
    k
    =\sum_{i=1}^k(\gamma_i^{-1})^{1/3}(\sqrt{\gamma_i})^{2/3}
    \le\left(\sum_{i=1}^k\gamma_i^{-1}\right)^{1/3}
        \left(\sum_{i=1}^k\sqrt{\gamma_i}\right)^{2/3}.
\]
Consequently, \(\operatorname{tr}(J_B^{-1})\ge k^3/S_B(p)^2\), proving~\eqref{eq:posterior-direction-error}.
This argument also applies when \(d=2\), for which \(k=1\).

\section{Proofs for Section~\ref{sec:finite-upper}}
\label{app:finite-estimation}

\subsection{Proof of Lemma~\ref{lem:soft-qme}}
\label{app:soft-qme}

Lemma~\ref{lem:soft-qme} builds on the phase and probability estimation methods of~\citet[Sections~6.2--6.4]{vanapeldoorn2023tomography}.
Throughout this subsection, \(d_{\mathbb T}\) denotes distance on the circle \(\mathbb T=\mathbb R/(2\pi\mathbb Z)\).

\paragraph{Phase estimation.}
Fix a power of two \(M\ge32\) and an integer \(m\ge1\).
We use the infinite precision version of~\citet[Algorithm~3]{vanapeldoorn2023tomography} with Fourier register dimension \(M\) and \(2m\) repetitions.
Let \(A_1,\ldots,A_{2m}\) be its independent randomized phase estimates at eigenphase \(\phi\).
For each \(j\), let \(d_j\) be the \(m\)-th smallest distance in \(\{d_{\mathbb T}(A_j,A_k):k\ne j\}\), and define
\begin{equation}
\label{eq:soft-selection-weights}
    w_j:=\exp\left(-\frac{mM}{\pi}d_j\right),
    \qquad
    p_j:=\frac{w_j}{\sum_{k=1}^{2m}w_k}.
\end{equation}
The algorithm samples an index \(I\) with probabilities \((p_j)_{j=1}^{2m}\) and returns \(\Phi_\infty:=A_I\).
By~\citet[Theorem~6.1]{vanapeldoorn2023tomography}, each \(A_j-\phi\) has a law independent of \(\phi\) and symmetric about zero on \(\mathbb T\).
The selection rule depends only on circular distances and therefore preserves both properties.
Thus \(\Phi_\infty=\phi+E\) modulo \(2\pi\), where the law of \(E\) is independent of \(\phi\) and symmetric about zero.
The proof of~\citet[Theorem~6.4, Eq.~(6.10)]{vanapeldoorn2023tomography} gives
\begin{equation}
\label{eq:soft-ideal-phase-tail}
    \Pr[d_{\mathbb T}(\Phi_\infty,\phi)>4\pi/M]
    \le2e^{-m/4}.
\end{equation}
Writing \(\lambda:=\mathbb E\cos E\), symmetry gives \(\mathbb E e^{i\Phi_\infty}=\lambda e^{i\phi}\).
To correct the bias from this factor, the probability estimator of~\citet[Section~6.3]{vanapeldoorn2023tomography} divides the cosine estimate by \(\lambda\).
Our implementation approximates this correction using independent calibration outputs at eigenphase zero.

Let \(\Phi_n\) denote the output when each random shift \(u\in[0,1)\) is replaced by \(u_n:=2^{-n}\lfloor2^nu\rfloor\) in both the circuit and the reported angle.
Each inverse Fourier transform has operator-norm error at most \(\beta\), and the classical sampler has total variation error at most \(\gamma\) from the probabilities in~\eqref{eq:soft-selection-weights} evaluated on the finite precision angles, uniformly over all input tables.
The following lemma controls how these approximations affect the output.

\begin{lemma}[Finite precision coupling]
\label{lem:soft-phase-coupling}
For every eigenphase \(\phi\), there is a coupling of \(\Phi_n\) and \(\Phi_\infty\) such that \(\Pr[d_{\mathbb T}(\Phi_n,\Phi_\infty)>h]\le q\), where
\begin{equation}
\label{eq:soft-coupling-budget}
    h:=\frac{2\pi}{M2^n},
    \qquad
    q:=2m(2\pi2^{-n}+\beta)
        +\frac{e^{8m2^{-n}}-1}{2}+\gamma.
\end{equation}
Consequently, every \(1\)-Lipschitz function \(f:\mathbb T\to\mathbb C\) with \(|f|\le1\) satisfies
\[
    \sup_\phi\left|\mathbb E_\phi f(\Phi_n)-\mathbb E_\phi f(\Phi_\infty)\right|
    \le h+2q.
\]
\end{lemma}

\begin{proof}
For a single phase estimation run, couple the random shifts by using \(u\) in the ideal circuit and \(u_n\) in the finite precision circuit.
Conditioned on \(u\), truncating the diagonal phase shift changes the circuit in operator norm by at most
\[
    \max_{0\le k<M}\left|e^{-2\pi iku/M}-e^{-2\pi iku_n/M}\right|
    \le2\pi2^{-n}.
\]
Including the inverse Fourier transform error, the two states before measurement differ in norm by at most \(2\pi2^{-n}+\beta\).
Their measured integer distributions therefore have total variation distance at most \(2\pi2^{-n}+\beta\).
A maximal coupling makes these integers agree except with this probability.
When they agree at an integer \(k\), the reported angles are \(2\pi(k+u)/M\) and \(2\pi(k+u_n)/M\) modulo \(2\pi\), so their circular distance is at most \(h\).
Apply this coupling independently to the \(2m\) runs at the fixed eigenphase \(\phi\).
All corresponding angles \(A_j,A'_j\) are within distance \(h\) except with probability \(2m(2\pi2^{-n}+\beta)\).

When all paired angles are within distance \(h\), every pairwise distance changes by at most \(2h\), so its \(m\)-th order statistic satisfies \(|d'_j-d_j|\le2h\).
Let \(w'_j,p'_j\) be the weights and probabilities in~\eqref{eq:soft-selection-weights} evaluated on \((A'_j)_{j=1}^{2m}\).
The exponential weights satisfy \(|\log(w'_j/w_j)|\le(mM/\pi)2h=4m2^{-n}\).
Their sum changes by at most the same multiplicative factors, giving
\[
    \left|\log\frac{p'_j}{p_j}\right|\le8m2^{-n},
    \qquad
    \frac12\sum_{j=1}^{2m}|p'_j-p_j|
    \le\frac{e^{8m2^{-n}}-1}{2}.
\]
The classical sampling error adds at most \(\gamma\) to this total variation bound.
Conditioned on the two angle tables, couple the selected indices maximally.
If the indices agree and all paired angles are within distance \(h\), the selected outputs are also within distance \(h\).
Adding the failure probabilities proves the claimed coupling bound.
On its success event, the difference between the two values of \(f\) is at most \(h\), and on its complement it is at most \(2\).
Taking expectations proves the moment bound, uniformly in \(\phi\).
\end{proof}

For \(\xi\in(0,1/8]\), choose
{
\begin{equation}
\label{eq:soft-finite-precision}
    \beta=\frac{\xi}{1000(m+1)},
    \qquad
    n=\left\lceil\log_2\frac{100}{\beta}\right\rceil,
    \qquad
    \gamma\le\frac{\xi}{100}.
\end{equation}
}
These parameters give \(2^{-n}\le\beta/100\) and \(8m2^{-n}\le\xi/12500\le1\).
Using \(e^z-1\le2z\) for \(0\le z\le1\), we obtain
\[
    q\le\left(\frac{4\pi+8}{10^5}+\frac{2}{1000}+\frac{1}{100}\right)\xi
    <\frac\xi4.
\]
Lemma~\ref{lem:soft-phase-coupling} therefore gives
\begin{equation}
\label{eq:soft-angular-coupling}
    \Pr\left[d_{\mathbb T}(\Phi_n,\Phi_\infty)>
        \frac{2\pi}{M2^n}\right]\le\frac\xi4.
\end{equation}
Also, \(h\le\pi\xi/(10^5M)\), so the moment bound in Lemma~\ref{lem:soft-phase-coupling} is at most \(\xi\).
Applying this inequality to \(f(a)=e^{ia}\) and \(g(a):=\min\{1-\cos a,1/4\}\) yields the two moment bounds used in calibration,
{
\begin{align}
    \sup_\phi\left|
        \mathbb E_\phi e^{i\Phi_n}-\lambda e^{i\phi}
    \right|&\le\xi,
    \label{eq:soft-finite-harmonic}\\
    \left|\mathbb E g(\Phi_n(0))-\mathbb E g(E)\right|
    &\le\xi.
    \label{eq:soft-calibration-moment}
\end{align}
}

\paragraph{Implementation with a reward oracle.}
The standard amplitude estimation construction~\citep{brassard2002quantum,montanaro2015quantum} prepares a state \(|\psi\rangle\) using one reward-oracle query and defines a Grover operator \(G\) whose eigenphases on this state satisfy \(\cos\phi=1-2\mu\).
By~\citet[Remark~2.5]{liu2026quantum}, each controlled application of \(G\) uses one ordinary query to \(\mathcal O\) and one to \(\mathcal O^\dagger\).

We next verify that the phase estimation guarantees apply when the target state is prepared only once.
Prepare \(|\psi\rangle\) once and perform the \(2m\) phase estimation runs sequentially on this same target, using a fresh phase register in every run.
For a fixed random shift and phase-register outcome \(y\), the induced Kraus operator on the target is a function of \(G\).
Writing \(G=\sum_\sigma e^{i\phi_\sigma}P_\sigma\), it has the form \(K_y=\sum_\sigma k_y(\phi_\sigma)P_\sigma\).
The Fourier gates act only on the phase register, so this spectral form also holds for an approximate Fourier transform.
For fixed random shifts, the full sequence of phase-register outcomes has probability
\begin{equation}
\label{eq:soft-shared-target}
    \Pr(y_1,\ldots,y_{2m})
    =\sum_\sigma\|P_\sigma\psi\|^2
        \prod_{j=1}^{2m}|k_{y_j}(\phi_\sigma)|^2.
\end{equation}
After averaging over the independent random shifts, the joint law remains a mixture of product distributions, with a common eigenphase in each component.
Since the preceding bounds hold uniformly in the eigenphase, they also apply to this mixture.

\paragraph{The mean estimator.}
We approximate \(1/\lambda\) by a truncated geometric series in \(1-\lambda\), using independent outputs at eigenphase zero to estimate its terms.
To bound both the bias and the failure probability by \(\eta\), set
\begin{equation}
\label{eq:soft-scalar-parameters}
    J=\left\lceil\log_4\frac{512}{\eta}\right\rceil,
    \qquad
    m=\left\lceil4\log\frac{512(J+1)}{\eta}\right\rceil,
    \qquad
    \xi=\frac{\eta}{1024(J+1)},
\end{equation}
and choose the finite precision parameters from~\eqref{eq:soft-finite-precision}.
Let \(\Phi_n\) be the output from the unknown reward oracle.
Independently generate \(J\) calibration outputs \(\Phi_n^{(j)}(0)\) by running the same implementation at eigenphase zero, and define
\begin{equation}
\label{eq:soft-reciprocal}
    X_j:=g(\Phi_n^{(j)}(0)),
    \qquad
    H_J:=\sum_{k=0}^J\prod_{j=1}^kX_j,
    \qquad
    Z:=\frac{1-H_J\cos\Phi_n}{2}.
\end{equation}
The empty product is \(1\).
Calibration uses known circuits with eigenphase zero, so it requires no reward-oracle queries.
Since \(X_j\in[0,1/4]\), we have \(1\le H_J\le4/3\) and \(Z\in[-1/6,7/6]\).

We first bound the bias of \(Z\), treating the arithmetic in~\eqref{eq:soft-reciprocal} as exact.
Let \(\rho=\mathbb E X_j\).
On the event \(d_{\mathbb T}(E,0)\le4\pi/M\), the clipping in \(g(E)\) is inactive because \(M\ge32\).
Equations~\eqref{eq:soft-ideal-phase-tail} and~\eqref{eq:soft-calibration-moment} thus give \(|(1-\rho)-\lambda|\le4e^{-m/4}+\xi\).
Independence of the calibration samples implies \(\mathbb EH_J=\sum_{k=0}^J\rho^k\), and hence \(|1-\lambda\mathbb EH_J|\le4^{-(J+1)}+\frac43(4e^{-m/4}+\xi)\).
Using~\eqref{eq:soft-finite-harmonic}, the independence of \(H_J\) and \(\Phi_n\), and \(\mu=(1-\cos\phi)/2\), we obtain
\begin{equation}
\label{eq:soft-scalar-bias}
    |\mathbb EZ-\mu|
    \le\frac12\,4^{-(J+1)}
        +\frac23(4e^{-m/4}+2\xi).
\end{equation}
This bound holds uniformly for each eigenphase and therefore for their mixture in~\eqref{eq:soft-shared-target}.

For the tail bound, equations~\eqref{eq:soft-ideal-phase-tail} and~\eqref{eq:soft-angular-coupling} show that each phase estimate, whether from the reward oracle or a calibration circuit, lies within \(5\pi/M\) of its eigenphase except with probability \(2e^{-m/4}+\xi/4\).
All \(J+1\) outputs satisfy these bounds except with probability \((J+1)(2e^{-m/4}+\xi/4)\).
On this event, each \(X_j\) is at most \(a_M:=1-\cos(5\pi/M)\le25\pi^2/(2M^2)<1/8\).
Thus \(H_J-1\le a_M/(1-a_M)\), which gives
\begin{equation}
\label{eq:soft-scalar-good-error}
    |Z-\mu|
    \le\frac{5\pi}{2M}+\frac{a_M}{2(1-a_M)}
    \le\frac{5\pi}{2M}+\frac{50\pi^2}{7M^2}
    <\frac{12}{M}.
\end{equation}

It remains to control the arithmetic error in the calibrated output.
Choose \(\kappa\le\xi/100\) and \(\kappa\le1/(100M^2)\).
Approximate each \(X_j\) to error at most \(\kappa\) and clip the approximation to \([0,1/4]\).
Evaluate \(H_J\) in reverse order by setting \(H^{(J+1)}=1\) and \(H^{(j)}=1+X_jH^{(j+1)}\) for \(j=J,\ldots,1\), so that \(H^{(1)}=H_J\) as defined in~\eqref{eq:soft-reciprocal}.
Evaluate each recursive update with total rounding error at most \(\kappa\), and clip its output to \([1,4/3]\).
Writing \(\widetilde H^{(j)}\) for the computed approximation, we have
\[
    |\widetilde H^{(j)}-H^{(j)}|
    \le\frac14|\widetilde H^{(j+1)}-H^{(j+1)}|+\frac73\kappa
    \le\frac{28}{9}\kappa.
\]
Approximate \(\cos\Phi_n\) to error at most \(\kappa\), clip it to \([-1,1]\), and round the final scalar output to error at most \(\kappa\).
After clipping to \([-1/6,7/6]\), the implemented output \(\widehat\mu\) satisfies \(|\widehat\mu-Z|\le(14/9+2/3+1)\kappa<4\kappa\).
The parameter choices in~\eqref{eq:soft-scalar-parameters} give
\[
    \frac12\,4^{-(J+1)}
        +\frac23(4e^{-m/4}+2\xi)+4\kappa<\eta,
    \qquad
    (J+1)(2e^{-m/4}+\xi/4)<\eta.
\]
Also, \(5\pi/(2M)+50\pi^2/(7M^2)+4\kappa<12/M\) for every \(M\ge32\).
Together with~\eqref{eq:soft-scalar-bias} and~\eqref{eq:soft-scalar-good-error}, these inequalities prove~\eqref{eq:soft-scalar-guarantee}.

Finally, preparing the target uses one reward-oracle query.
Each of the \(2m\) phase estimation runs uses \(M-1\) controlled Grover iterates, each costing two queries.
With \(m=O(\log(1/\eta))\) from~\eqref{eq:soft-scalar-parameters}, the total is \(1+4m(M-1)=O(M\log(1/\eta))\).
The circuit lengths and numerical precisions are fixed before the call, so the query bound is deterministic.

\subsection{Proof of Lemma~\ref{lem:soft-design}}
\label{app:soft-design-proof}

\begin{proof}
We separate the estimation error into its conditional bias and its fluctuation around the conditional mean.
For each \(a\in S\), let \(\mathcal G_a:=\{|\widehat\mu_a-\mu(a)|\le12/M_a\}\), and set \(\mathcal G:=\bigcap_{a\in S}\mathcal G_a\).
By Lemma~\ref{lem:soft-qme}, \(\Pr(\mathcal G^c)\le|S|\eta\le c_s r\eta\le\delta/4\).
For the bias, write \(b_a:=\mathbb E[\widehat\mu_a\mid\mathcal G_a]\).
The bias and range bounds in Lemma~\ref{lem:soft-qme} give
\[
    |b_a-\mu(a)|
    \le\frac{\eta+(7/6)\Pr(\mathcal G_a^c)}
        {1-\Pr(\mathcal G_a^c)}
    \le3\eta.
\]
Consequently, for every \(x\in\mathcal X\),
\begin{equation}
\label{eq:soft-design-conditional-bias}
    \left|\sum_{a\in S}\alpha_a(x)(b_a-\mu(a))\right|
    \le3\eta\sqrt{C_G r}\le\epsilon/4.
\end{equation}

Conditional on \(\mathcal G\), the mean estimates \(\widehat\mu_a\) remain independent because each event \(\mathcal G_a\) depends only on \(\widehat\mu_a\).
Their conditional means are \(b_a\), and their interval lengths are at most \(24/M_a\), so Hoeffding's inequality gives
\begin{align}
    \Pr\left[
        \left|\sum_{a\in S}\alpha_a(x)(\widehat\mu_a-b_a)\right|
        >\frac\epsilon2
        \,\middle|\,\mathcal G
    \right]\le
    2\exp\left(-\frac{\epsilon^2}
        {1152\sum_{a\in S}\alpha_a(x)^2/M_a^2}\right).
    \label{eq:soft-design-hoeffding}
\end{align}
The choice of \(M_a\) and~\eqref{eq:design-coefficients} give
\[
    \sum_{a\in S}\frac{\alpha_a(x)^2}{M_a^2}
    \le
    \frac{\epsilon^2}{A_0^2r\log(8|\mathcal X|/\delta)}
    \sum_{a\in S}\frac{\alpha_a(x)^2}{\pi(a)}
    \le
    \frac{C_G\epsilon^2}{A_0^2\log(8|\mathcal X|/\delta)}.
\]
The probability in~\eqref{eq:soft-design-hoeffding} is therefore at most \(\delta/(4|\mathcal X|)\).
A union bound over \(\mathcal X\), together with~\eqref{eq:soft-design-conditional-bias} and the bound on \(\Pr(\mathcal G^c)\), proves simultaneous accuracy.

For the query bound, rounding \(M_a\) to a power of two increases it by at most a factor of two.
Since \(\sum_{a\in S}\sqrt{\pi(a)}\le\sqrt{|S|}\), we have
\[
    \sum_{a\in S}M_a
    \le C|S|+
        C\frac{\sqrt{r\log(8|\mathcal X|/\delta)}}{\epsilon}
        \sum_{a\in S}\sqrt{\pi(a)}
    \le C\left(r+
        \frac{r\sqrt{\log(8|\mathcal X|/\delta)}}{\epsilon}\right).
\]
Multiplying by the \(O(\log(1/\eta))\) factor from Lemma~\ref{lem:soft-qme} and substituting the choice of \(\eta\) gives~\eqref{eq:soft-design-cost}.
\end{proof}

\subsection{The \QGElim Algorithm}
\label{app:estimation-budgets}

{
\begin{algorithm}[H]
\caption{\QGElim}
\label{alg:combined-elimination}
\KwIn{Finite action set \(\mathcal A\), horizon \(T\ge2\), confidence parameter \(\delta\in(0,1)\)}
\(L_T\leftarrow\lceil\log_2T\rceil\), \(\mathcal A_0\leftarrow\mathcal A\),
    \(\delta'\leftarrow\delta/(L_T+1)\)\;
\For{\(\ell=0,\ldots,L_T\)}{
    \(\epsilon_\ell\leftarrow2^{-\ell}\)\;
    \If{\(|\mathcal A_\ell|=1\)}{
        Select the remaining action until time \(T\) and stop\;
    }
    Compute a design \((S_\ell,\pi_\ell,V_\ell)\) for \(\mathcal A_\ell\) using Lemma~\ref{lem:small-support-design}\;
    Compute \(Q_{\rm soft}\) and \(Q_{\rm QMC}\) from~\eqref{eq:combined-query-budgets-soft} and~\eqref{eq:combined-query-budgets-qmc}
        for inputs \((\mathcal A_\ell,S_\ell,\pi_\ell,V_\ell,\epsilon_\ell,\delta')\)\;
    \eIf{\(Q_{\rm soft}\le Q_{\rm QMC}\)}{
        \(\{\widehat\mu_\ell(x)\}_{x\in\mathcal A_\ell}
            \leftarrow\SoftDesignEstimate
            (\mathcal A_\ell,S_\ell,\pi_\ell,V_\ell,\epsilon_\ell,\delta')\)\;
    }{
        \(\{\widehat\mu_\ell(x)\}_{x\in\mathcal A_\ell}
            \leftarrow\QMCDesignEstimate
            (\mathcal A_\ell,S_\ell,\pi_\ell,V_\ell,\epsilon_\ell,\delta')\)\;
    }
    \(\widehat\mu_\ell^\star\leftarrow
        \max_{x\in\mathcal A_\ell}\widehat\mu_\ell(x)\)\;
    \(\mathcal A_{\ell+1}\leftarrow
        \{x\in\mathcal A_\ell:
        \widehat\mu_\ell(x)\ge\widehat\mu_\ell^\star-2\epsilon_\ell\}\)\;
}
Select any action in \(\mathcal A_{L_T+1}\) for all remaining rounds\;
\end{algorithm}
}

\end{document}